\documentclass[10pt,twocolumn,letterpaper]{article}

\usepackage{cvpr}              
\usepackage{multirow}
\usepackage[dvipsnames]{xcolor}

\usepackage{etoc}

\newcommand{\redcross}{{\color{red}$\times$}}
\newcommand{\greencheck}{{\color{Green} \checkmark}}
\usepackage[section]{placeins}
\newcommand\blfootnote[1]{%
  \begingroup
  \renewcommand\thefootnote{}\footnote{#1}%
  \addtocounter{footnote}{-1}%
  \endgroup
}
\definecolor{cvprblue}{rgb}{0.21,0.49,0.74}
\usepackage[pagebackref,breaklinks,colorlinks,citecolor=cvprblue]{hyperref}
\newcommand{\ModelName}{\emph{PhysHOI}\xspace}

\def\paperID{5597} 
\def\confName{CVPR}
\def\confYear{2024}

\title{PhysHOI: Physics-Based Imitation of Dynamic Human-Object Interaction}

\author{Yinhuai~Wang\textsuperscript{\rm 1,2\S}\quad
Jing~Lin\textsuperscript{\rm 3}\quad
Ailing~Zeng\textsuperscript{\rm 2}$^{\dagger}$\quad
Zhengyi~Luo\textsuperscript{\rm 4}\quad
Jian~Zhang\textsuperscript{\rm 1}$^{\dagger}$\quad
Lei~Zhang\textsuperscript{\rm 2}
\\
\normalsize{
\textsuperscript{\rm 1}Peking University\quad \textsuperscript{\rm 2}International Digital Economy Academy\quad
\textsuperscript{\rm 3}Tsinghua University\quad
\textsuperscript{\rm 4}Carnegie Mellon University
}
\\
}

\begin{document}

\twocolumn[{
\renewcommand\twocolumn[1][]{#1}
\vspace{-0.3cm}
\maketitle
\centering
\vspace{-0.5cm}
\includegraphics[width=\textwidth]{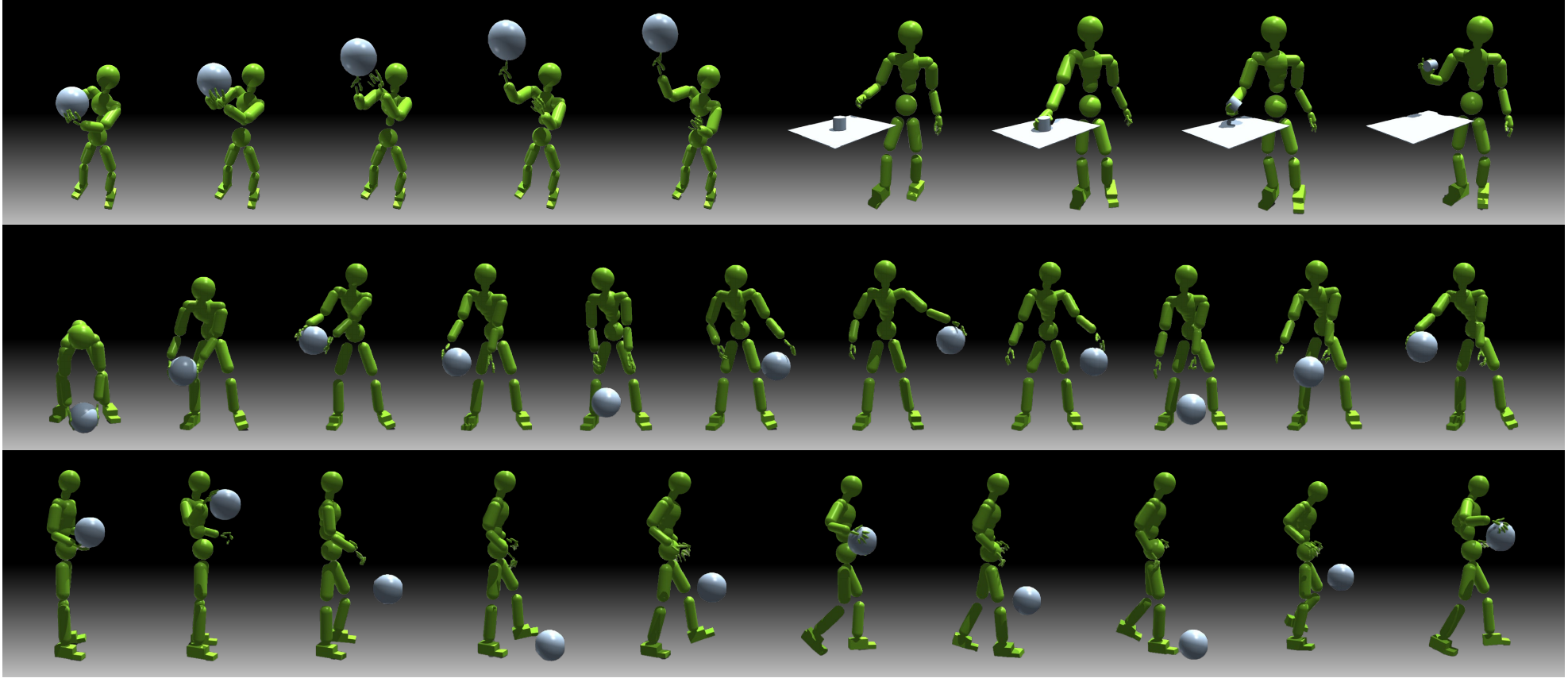}
\vspace{-0.5cm}
\captionsetup{type=figure}
\caption{Our method controls physically simulated humanoids to perform various dynamic interaction skills. Our method can imitate human-object interaction across dynamic scenarios without task-specific rewards. \textbf{Top-to-bottom}: Fingertip spin basketball (\textit{left}), Grasp (\textit{right}); Pick up and dribble; Walk while dribbling. Project page and video demonstrations: \textcolor{blue}{https://wyhuai.github.io/physhoi-page/}
}
\label{fig: teaser}
\vspace{0.2cm}
}]
\etocdepthtag.toc{mtchapter}

\begin{abstract}
\vspace{-0.3cm}
Humans interact with objects all the time. Enabling a humanoid to learn human-object interaction (HOI) is a key step for future smart animation and intelligent robotics systems. However, recent progress in physics-based HOI requires carefully designed task-specific rewards, making the system unscalable and labor-intensive. This work focuses on dynamic HOI imitation: teaching humanoid dynamic interaction skills through imitating kinematic HOI demonstrations. It is quite challenging because of the complexity of the interaction between body parts and objects and the lack of dynamic HOI data.
To handle the above issues, we present PhysHOI, the first physics-based whole-body HOI imitation approach without task-specific reward designs. Except for the kinematic HOI representations of humans and objects, we introduce the contact graph to model the contact relations between body parts and objects explicitly. A contact graph reward is also designed, which proved to be critical for precise HOI imitation. Based on the key designs, PhysHOI can imitate diverse HOI tasks simply yet effectively without prior knowledge. To make up for the lack of dynamic HOI scenarios in this area, we introduce the \emph{BallPlay} dataset that contains eight whole-body basketball skills. We validate PhysHOI on diverse HOI tasks, including whole-body grasping and basketball skills.
\vspace{0pt}
\blfootnote{$\S$Work done during an internship at IDEA; $^{\dagger}$~Corresponding authors.}
\end{abstract}

\vspace{-0.3cm}
\section{Introduction}
\label{sec:intro}
Creating realistic and agile interactions between objects and humanoids is a long-standing and challenging problem in computer animation, robotics, and human-computer interaction (HCI)~\cite{luck2000applying,stacey2012animation,geijtenbeek2012interactive,mourot2022survey}. 
Recently, due to advances in deep reinforcement learning and physics simulation, methods that learn humanoid control from demonstrations (e.g., kinematic motions from motion capture data) have been widely used in animation \cite{RoboImitationPeng20,DeepMimic,mvae,amp,ase,won2022physics,tennis} and robotics \cite{escontrela2022adversarial,bohez2022imitate,TWiRLSmith2023, bahl2022human}. However, most existing physics-based imitation learning methods focus on mimicking isolated human motion without considering human-object interactions, especially for whole-body (i.e., full-body with hands) motions, as involving articulated fingers and complex object dynamics can be exceedingly challenging. 

There exist HOI tasks conditioned on goals, in which the humanoid is tasked with manipulating objects to reach certain kinematic states \cite{pmp, InterPhysHassan2023, tennis,braun2023physically}. Although they perform well on specific tasks, e.g., playing tennis \cite{tennis}, these methods may find the learning paradigm difficult to generalize to new types of interaction and skills, considering each needs specially designed task rewards. For example, basketball games consist of multiple basic skills: dribbling, shooting, passing, fingertip spinning, etc. It is extremely laborious to design task-specific rewards for all skills.

To solve diverse HOI tasks within a unified solution, we resort to the HOI imitation, i.e., recreating diverse HOI skills in simulation from kinematic demonstrations. It should be noted that the difference between HOI imitation and previous imitation methods \cite{DeepMimic,amp} is that they imitate isolated human motion, while our aim is to recreate motion for both the human and the object. During simulation, the object is passive and can only be controlled indirectly by the humanoid, making this task extremely challenging.  

Correspondingly, we introduce the first physics-based whole-body HOI imitation framework, called \ModelName, which makes HOI imitation feasible in diverse scenarios. Given a kinematic HOI demonstration sequence, \ModelName can imitate the HOI skill without task-specific knowledge and reward designs. Considering the importance of contact in describing HOI semantics, we introduce a novel general-purpose Contact Graph (CG), where nodes are whole-body humanoid body segments and objects, and each edge denotes the binary contact information between two nodes, to enhance the commonly-used kinematic HOI representation (e.g., the human motion, object motion, and interaction graph \cite{ig2023}). Based on the contact-aware HOI representation, we present a task-agnostic HOI imitation reward that multiplies the kinematic rewards with the proposed contact graph reward (CGR), which effectively eliminates local optima in kinematic-only rewards. To make up for the lack of dynamic HOI data, we introduce the \emph{BallPlay} dataset, which contains eight diverse human-basketball interaction demonstrations with high-quality SMPL-X~\cite{smplx} and object motions, and contact labels.

\begin{figure*}[t]
  \centering
  \vspace{-0.3cm}
  \includegraphics[width=1\linewidth]{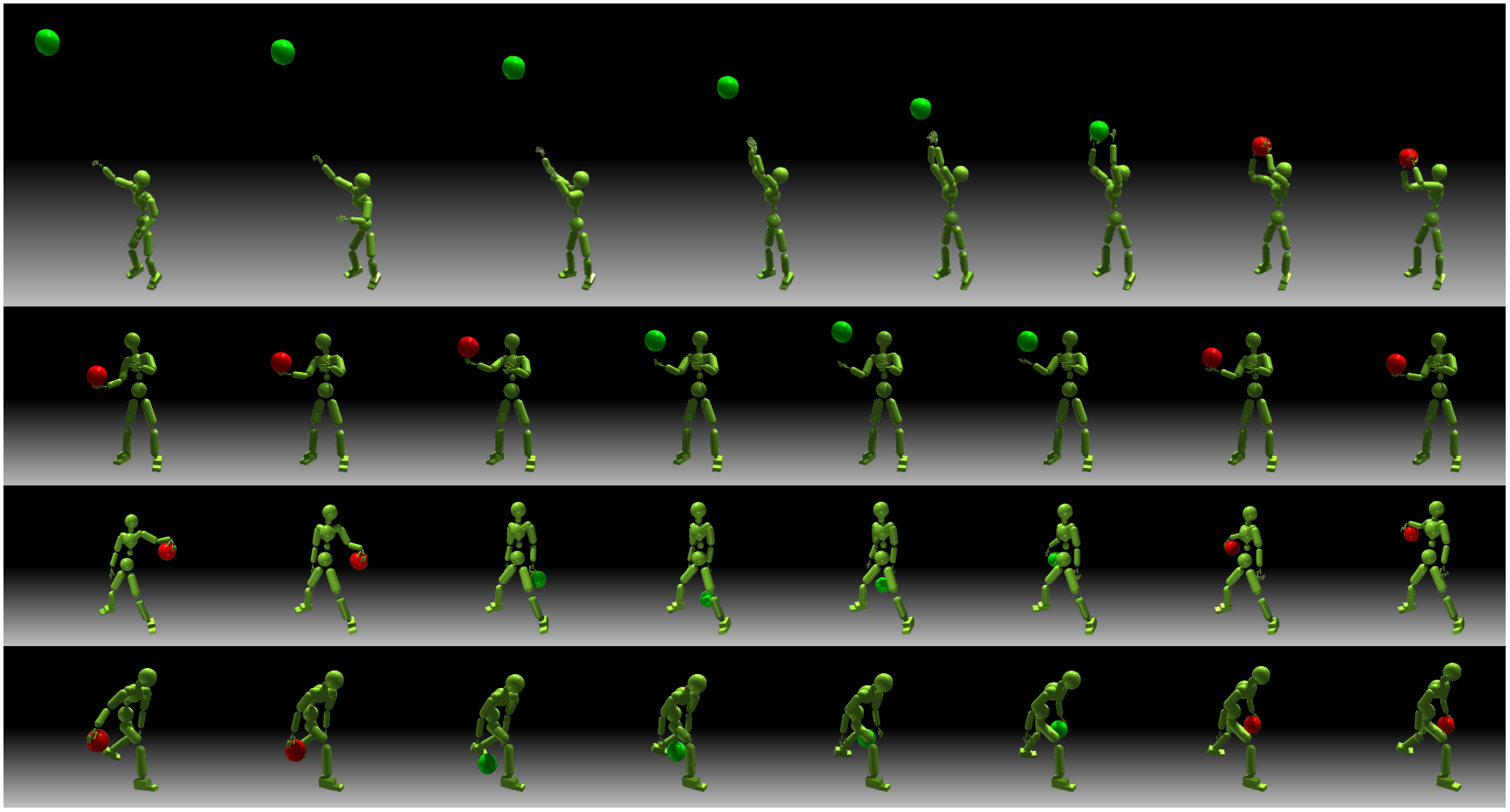}
  \vspace{-0.4cm}
  \caption{Our method controls simulated humanoids to perform various basketball skills. \textbf{Top-to-bottom}: 1) Rebound; 2) Single-hand toss and catch; 3) Back dribbling; 4) Cross-leg dribble. We mark \textcolor{red}{red} when the object has contact with the humanoid. }
  \vspace{-0.2cm}
\label{fig: more results} 
\end{figure*}

Our core contributions can be summarized below:

\begin{itemize}
    \item Aiming at learning dynamic HOI skills with the whole-body humanoids in simulation, we present \ModelName. This first whole-body HOI imitation approach learns interaction skills directly from kinematic HOI demonstrations.

    \item To make the method task-agnostic towards general HOI learning, we introduce a general-purpose contact graph (CG) as a critical complement. Correspondingly, we design a novel contact graph reward (CGR), which effectively guides whole-body humanoids to manipulate objects or interact with high-dynamic objects precisely without task-specific reward designs.
    
    \item We introduce the \emph{BallPlay} dataset to fill the gap in missing dynamic HOI datasets.   
\end{itemize}
We validate the effectiveness of our proposed method comprehensively on five whole-body grasping cases and eight basketball skills, as shown in Fig.~\ref{fig: teaser} and Fig.~\ref{fig: more results}. Compared with previous state-of-the-art methods, \ModelName significantly improves the success rate and object trajectory errors. \ModelName is simple yet effective and easy to generalize across diverse types of HOI tasks. We hope this work could pave the path for humanoids to learn general HOI skills.

\begin{table}
\centering
\begin{center}
\resizebox{0.5\textwidth}{!}{%
\begin{tabular}{ll|cccc}
   \toprule
   & & \textbf{{High-Dynamic}}  & \textbf{{Contact}}  & \textbf{{Task-Agnostic}}  & \textbf{Whole-Body} \\
     \textbf{Method} &\textbf{Year}     &   \textbf{HOI} &  \textbf{Guidance} & \textbf{Rewards} &   \textbf{Motions}\\
    \midrule
    DeepMimic \cite{DeepMimic} &2018   & \redcross    & \redcross                & \redcross & \redcross    \\
    AMP \cite{amp} &2021   & \redcross    & \redcross                & \redcross & \redcross    \\
    Hassan \etal \cite{InterPhysHassan2023}  &2023     & \redcross & \redcross               & \redcross   & \redcross   \\
    Vid2tennis \cite{tennis} &2023   & \greencheck    & \greencheck                & \redcross & \redcross    \\
    PMP \cite{pmp}  &2023  & \redcross    & \redcross                 & \redcross & \greencheck    \\
    Zhang \etal \cite{ig2023}   &2023   & \redcross  & \redcross                & \greencheck & \redcross \\
    Braun \etal \cite{braun2023physically}&2023 & \redcross & \greencheck             & \redcross   & \greencheck   \\
    
    \midrule
    \textbf{\ModelName (Ours)} & 2023& \greencheck & \greencheck         & \greencheck & \greencheck  \\
   \bottomrule
\end{tabular}
}
\end{center}
\vspace{-0.4cm}
\caption{
Comparisons of the most related \textbf{physics-based} methods.}
\label{tab: method comparison}
\vspace{-0.4cm}
\end{table}

\section{Related Work}
Related works on physically simulated humanoids can be divided into isolated motion imitation and human-object interaction learning. We compare the most related \textbf{physics-based} work in Tab.~\ref{tab: method comparison} and present the related \textbf{kinematics-based} methods in the Appendix.

\noindent{\textbf{Physics-Based Isolated Motion Imitation.}}
Physics-based methods generate motions via motor control in physics simulation \cite{makoviychuk2021isaac, todorov2012mujoco, coumans2016pybullet}, which addresses the physically implausible artifacts of kinematic methods \cite{humanml3d, mdm, motiondiffuse, zhang2023t2m, chen2023executing,humantomato}, \eg, penetration, sliding, and floating. Early works rely on hand-crafted \cite{Hodgins1995}, model-based \cite{Coros2010}, and optimization-based \cite{Tan_Gu_Liu_Turk_2014} controllers, which suffer from motion artifacts and complex parameter tuning. Recently, DeepMimic \cite{DeepMimic} proposes tracking motion capture sequences via imitation learning and deep reinforcement learning (RL). To make the generation diverse, AMP \cite{amp} uses generative adversarial imitation learning (GAIL) \cite{ho2016generative}, which learns the state transition distribution of unstructured motion dataset. Except for RL-based methods, some methods use physics simulation with fully differentiable pipelines~\cite{ren2023diffmimic,fussell2021supertrack,yao2022controlvae}. Some works combine kinematic methods with physics-based imitation policy \cite{yuan2022physdiff, shi2023controllable, luo2023perpetual, tennis} to obtain physically plausible motions. However, these methods only imitate isolated human motion.

\begin{figure*}[t]
  \centering
  \vspace{-0.2cm}
  \includegraphics[width=1\linewidth]{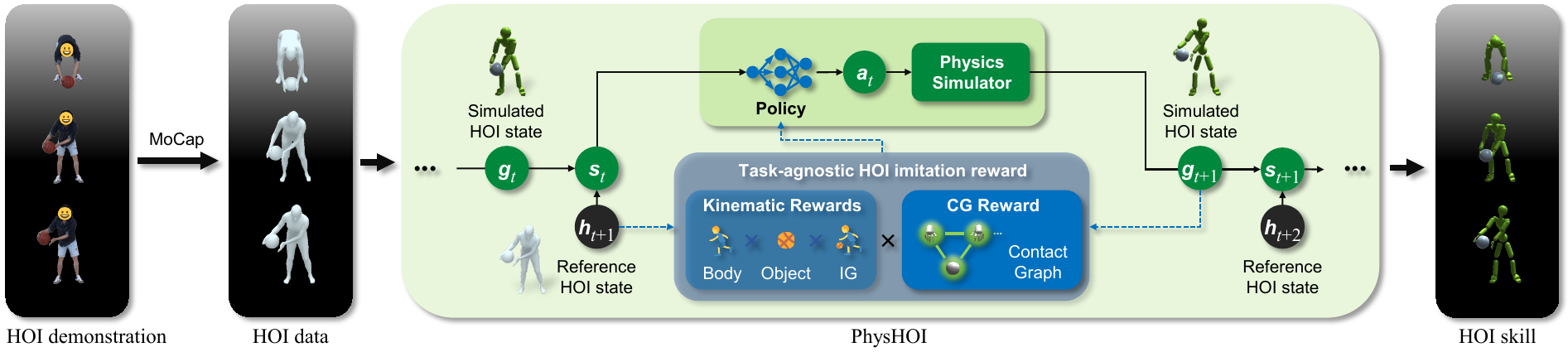} 
  \vspace{-0.4cm}
  \caption{\textbf{Framework overview}: The proposed pipeline of learning HOI skills from HOI demonstrations. 
  We can obtain kinematic HOI data using mocap devices or estimated from monocular videos. 
  Then, we transfer the HOI data into the reference HOI states, which is a contact-aware HOI representation \{\emph{human motion, object motion, interaction graph (IG), contact graph (CG)}\} for \ModelName to learn. 
  \textbf{\ModelName}: The training process of \ModelName consists of loops of simulation and optimization. Given the simulated HOI state $\boldsymbol{g}_t$ and reference HOI state $\boldsymbol{h}_{t+1}$, the policy outputs the action $\boldsymbol{a}_t$, then the simulated HOI state will be updated by the physics simulator. 
  For each time step, we calculate the proposed task-agnostic HOI imitation reward, including kinematic rewards and the key CG reward.
  We train the policy until converges, where it can control simulated humanoids to reproduce the reference HOI skills.} 
\label{fig: method} 
\vspace{-0.5cm}
\end{figure*}

\noindent{\textbf{Physics-based Human-Object Interaction.}}
Human-object interaction (HOI) faces more challenges and is less explored than isolated motion generation due to the complexity of human body parts, various objects, and diverse interactions. 
Compared to kinematics-based methods~\cite{starke2020local,xiao2023unified,pi2023hierarchical}, physics-based methods are especially advantageous for generating human-object interaction since the physics simulator can explicitly constrain the dynamics of humanoids and objects. Using simulation, one can create realistic human-object interactions such as carrying boxes \cite{InterPhysHassan2023, xie2023hierarchical}, striking \cite{ase}, climbing ropes \cite{pmp}, grasping \cite{braun2023physically}, and even challenging sports such as skating \cite{llbskating}, playing basketball \cite{llbasketball}, soccer~\cite{xie2022learning,hong2019physics}, football~\cite{liu2022motor} and tennis \cite{tennis}. However, these methods rely on manual task-specific reward designs, making them difficult to generalize. 
Besides, there are active topics on human-scene interaction \cite{xiao2023unified,pan2023synthesizing} and hand-object interaction \cite{garcia2020physics,christen2022d,patel2022learning,xu2023unidexgrasp,dasari2023learning,zhang2023artigrasp}. 

A highly related work is \cite{ig2023}, where the proposed interaction graph \cite{ho2010spatial} is integrated into kinematic rewards to learn the simulated multi-character interaction. Though effective in simple human-object interactions (\eg picking up boxes), the kinematic-only rewards are prone to fall into local optimal when dealing with high-dynamic scenarios. 
Another closely related work is \cite{braun2023physically}, a framework for physically plausible whole-body grasping. Similar to previous kinematics-based grasping synthesis methods \cite{wu2022saga,ghosh2022imos}, they design task-specific contact rewards as guidance for learning grasp tasks. However, their method contains multiple networks and training rounds with prior learning dedicated to the grasp tasks. Our work explores whole-body object interaction with static or high-dynamic objects (e.g., grasping, playing with balls) without designing task-specific rewards.

\section{Method}

\subsection{Preliminaries on Reinforcement Learning}
\label{sec:rl}
Our method is based on reinforcement learning, where the agent interacts with the environment according to a policy to maximize reward. At each time step $t$, the agent takes the system states $\mathbf{s}_{t}$ as input and outputs an action $\mathbf{a}_{t}$ by sampling the policy distribution $\boldsymbol{\pi}(\mathbf{a}_{t}|\mathbf{s}_{t})$. According to the physics simulator $\boldsymbol{f}(\mathbf{s}_{t+1}|\mathbf{a}_{t},\mathbf{s}_{t})$, the new action $\mathbf{a}_{t}$ will result in a new state $\mathbf{s}_{t+1}$. Then a reward $r_{t}=r(\mathbf{s}_{t}, \mathbf{a}_{t},\mathbf{s}_{t+1})$ can be calculated, and the goal is to learn a policy that maximizes the expected return $    \mathcal{R}(\boldsymbol{\pi}) = \mathbb{E}_{p_{\boldsymbol{\pi}}(\boldsymbol{\tau})}\left[ \sum_{t=0}^{T-1}\gamma^{t}r_{t}\right]$,
where $\boldsymbol{\tau}=\{\mathbf{s}_{0},\mathbf{a}_{0}, r_{0}, ..., \mathbf{s}_{T-1},\mathbf{a}_{T-1}, r_{T-1}, \mathbf{s}_{T}\}$ represents the trajectory, $p_{\boldsymbol{\pi}}(\boldsymbol{\tau})$ is the probability density function (PDF) of the trajectory. $T$ denotes the time horizon of a trajectory, and $\gamma\in [0, 1)$ is a discount factor. Following prior arts \cite{amp}, We use  PPO \citep{schulman2017proximal} to optimize the policy.

\subsection{Task Definition}
\label{sec:task define}
Given a reference demonstration of Human-Object Interaction (HOI), represented by a kinematic sequence of human and object states, HOI imitation aims to train a policy to control the simulated humanoid to reproduce the given HOI. Unlike kinematics-based methods \cite{starke2020local,ghosh2022imos,kapon2023mas}, the core challenge here is that the object is not directly controllable. Instead, we can only indirectly manipulate objects by controlling the humanoid. The whole-body humanoid follows the SMPL-X~\cite{smplx} kinematic tree and has a total of 52 body parts and 51$\times$3 DoF actuators where 30$\times$3 DoF is for the hands and 21$\times$3 DoF for the rest of the body. The object is represented as a rigid body with a simplified mesh for collision detection. Details on the HOI data preprocessing can be found in the Appendix.

\subsection{Overview of PhysHOI}
\label{sec:overview}

In Fig.~\ref{fig: method}, we present the overview of the proposed pipeline for HOI imitation.
The core of \ModelName is the general-purpose contact graph (Sec.~\ref{sec: cg}), introducing it into the kinematics-based HOI representation as the contact-aware HOI representation (Sec.~\ref{sec:hoi representation}), and the task-agnostic HOI imitation reward (Sec.~\ref{sec:reward}).
Similar to prior arts \cite{DeepMimic}, policy training consists of a loop of simulation and optimization: the policy is first used for simulation and then optimized through RL (Sec.~\ref{sec:rl}), then the updated policy continues simulation to collect more experiences and repeat this simulation-optimization loop until converges. 
Specifically, the state $\boldsymbol{s}_{t}$ (Sec.~\ref{sec:state}) consists of the simulated HOI state $\boldsymbol{g}_{t}$ and the reference HOI state $\hat{\boldsymbol{h}}_{t+1}$ (Sec.~\ref{sec:hoi representation}). The policy network (Sec.~\ref{sec: policy and action}) takes the state $\boldsymbol{s}_{t}$ as input and outputs an action $\boldsymbol{a}_{t}\in\mathbb{R}^{51\times 3}$. The PD controller outputs joint torques (omitted in Fig.~\ref{fig: method} for simplicity). Then the physics simulator calculates the updated simulated HOI state $\boldsymbol{g}_{t+1}$.

\vspace{-0.2cm}
\subsection{General-purpose Contact Graph}
\label{sec: cg}
Contact information is essential for HOI. Though contact forces are hard to acquire, binary contact labels are easy to extract from kinematic HOI data by calculating the mesh collision or distances. In addition, estimating the contact region from images is also a feasible direction \cite{tripathi2023deco}. 
We propose a general-purpose Contact Graph (CG) to enhance the HOI representation across diverse interaction imitations. 

\begin{figure}[h]
  \centering
  \vspace{-0.4cm}
  \includegraphics[width=1\linewidth]{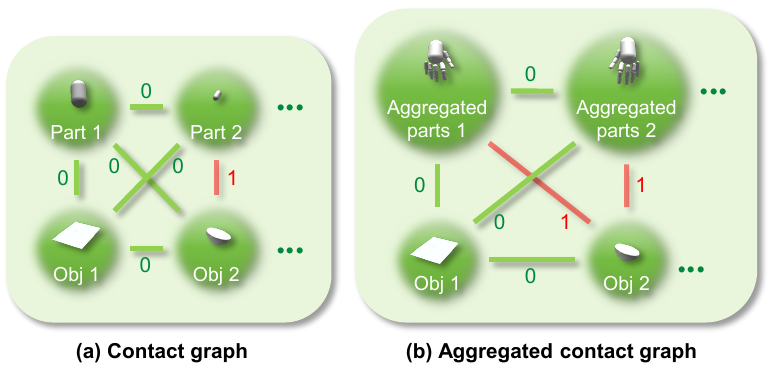}
  \vspace{-0.6cm}
  \caption{\textbf{Contact Graph}. (\textbf{a}) The nodes of the complete contact graph consist of all the objects and humanoid body parts. Each edge stores a binary contact label.  (\textbf{b}) A node in the aggregated contact graph can contain multiple body parts.}
  \vspace{-0.4cm}
\label{fig: cg} 
\end{figure}

\noindent{\textbf{Complete CG.}} As illustrated in Fig.~\ref{fig: cg} (a), the CG is a complete graph in which every pair of distinct nodes connected by a unique edge, defined as $\mathcal{G}=\{\mathcal{V},\mathcal{E}\}$, where $\mathcal{V}$ is the set of $k$ nodes and $\mathcal{E}\in\{0,1\}^{k(k-1)/2}$ is the set of edges. The nodes consist of all the objects and humanoid body parts. Each edge stores a binary label that denotes the contact between two nodes, where 1 represents contact, and 0 means no contact. Here, we only use the edge set of CG. The edge value is calculated frame by frame. For an HOI sequence, we can calculate a corresponding CG sequence $\{\mathcal{E}_t\}$, which explicitly describes the mutual contact relationship between objects and bodies at different moments. Considering the trend of hinged objects, objects here can also be broken up into finer parts \cite{fan2023arctic,geng2023gapartnet,zhang2023artigrasp}. 

\noindent{\textbf{Aggregated CG.}} 
There are three limitations in the implementation of the complete CG defined above. First, in the whole-body grasp scenario, the complete CG has 154 nodes (152 body parts, 1 table, and 1 object) and 11781 edges, which is costly in memory and computation. Second, many contact labels may be noisy due to inevitable errors from kinematic HOI estimation and annotation. Third, the contact APIs in existing physics simulation environments \cite{makoviychuk2021isaac} are not yet complete. To address these challenges, we propose the aggregated CG, where the graph node can be composed of multiple aggregated parts, as illustrated in Fig.~\ref{fig: cg} (b). For example, in basketball scenarios, we can aggregate two hands as one node, the rest of the bodies as a node, and the ball as a node. As long as there is any collision between two aggregated nodes, \eg, the fingertip touches the ball; the corresponding edge value becomes one. Contact between parts belonging to the same node is not considered. The aggregated CG significantly simplifies the use of CG and proves to be critical for simulated humanoid learning accurate HOI imitation.

\subsection{Contact-Aware HOI Representation}
\label{sec:hoi representation}
Zhang \etal \cite{ig2023} introduced kinematics-based representation, including human (subject) motion $\hat{\boldsymbol{s}}^{sbj}_{t}$, object motion $\hat{\boldsymbol{s}}^{obj}_{t}$, interaction graph (IG) $\hat{\boldsymbol{s}}^{ig}_{t}$. We further introduce the proposed CG $\hat{\boldsymbol{s}}^{cg}_{t}$ to represent the reference HOI state $\hat{\boldsymbol{h}}_{t}$:
\begin{equation}
\begin{aligned}
    \hat{\boldsymbol{h}}_{t} = \{\hat{\boldsymbol{s}}^{sbj}_{t},\hat{\boldsymbol{s}}^{obj}_{t},\hat{\boldsymbol{s}}^{ig}_{t},\hat{\boldsymbol{s}}^{cg}_{t} \}.
    \label{eq: h}
\end{aligned}
\end{equation}
The human motion $\hat{\boldsymbol{s}}^{sbj}_{t}$ consists of the position $\hat{\boldsymbol{s}}_{t}^{p}\in\mathbb{R}^{52\times 3}$, 6D rotation $\hat{\boldsymbol{s}}_{t}^{r}\in\mathbb{R}^{52\times 6}$, position velocity $\hat{\boldsymbol{s}}_{t}^{pv}\in\mathbb{R}^{52\times 3}$, and rotation velocity $\hat{\boldsymbol{s}}_{t}^{rv}\in\mathbb{R}^{52\times 6}$. The object motion consists of the object position $\hat{\boldsymbol{s}}_{t}^{op}\in\mathbb{R}^{1\times 3}$, object 6D rotation $\hat{\boldsymbol{s}}_{t}^{or}\in\mathbb{R}^{1\times 6}$, object position velocity $\hat{\boldsymbol{s}}_{t}^{opv}\in\mathbb{R}^{1\times 3}$, and object rotation velocity $\hat{\boldsymbol{s}}_{t}^{orv}\in\mathbb{R}^{1\times 6}$. All velocities here are calculated via discrete differences, \eg $\hat{\boldsymbol{s}}_{t}^{opv}=\hat{\boldsymbol{s}}_{t}^{op}-\hat{\boldsymbol{s}}_{t-1}^{op}$. The IG used in this paper is inspired by previous works \cite{ho2010spatial,ig2023}, but with a simpler definition: A set of positional vectors pointed from the objects to the contact bodies (the body parts that have been in contact with the object throughout the sequence). The IG $\hat{\boldsymbol{s}}^{ig}_{t}\in\mathbb{R}^{n\times m\times 3}$ can be seen as a relative position representation of the object, where $n$ denotes the contact body numbers and $m$ denotes the object numbers. The CG $\hat{\boldsymbol{s}}^{cg}_{t}\in\{0,1\}^{k(k-1)/2}$, where $k$ denotes the number of CG nodes.

\subsection{Task-agnostic HOI Imitation Reward}
\label{sec:reward}

Corresponding to the reference HOI state (Sec.~\ref{sec:hoi representation}), the proposed task-agnostic HOI imitation reward consists of four parts: the body motion reward $r_{t}^{b}$, the object motion reward $r_{t}^{o}$, the IG reward $r_{t}^{ig}$, and the CG reward $r_{t}^{cg}$. To obtain balanced reward values, we \emph{multiply} these rewards to request that none of them be small:
\begin{equation}
\begin{aligned}
    r_{t} = r_{t}^{b}*r_{t}^{o}*r_{t}^{ig}*r_{t}^{cg},
    \label{eq: r}
\end{aligned}
\end{equation}

\noindent{\textbf{Body Motion Reward}}
can be formulated as:
\begin{equation}
\begin{aligned}
    r_{t}^{b} = r_{t}^{p}*r_{t}^{r}*r_{t}^{pv}*r_{t}^{rv},
    \label{eq: rb}
\end{aligned}
\end{equation}
where $r_{t}^{p}$, $r_{t}^{r}$, $r_{t}^{pv}$, $r_{t}^{rv}$ are the humanoid position reward, rotation reward, position velocity reward, and rotation velocity reward, formulated below:
\begin{equation}
\begin{aligned}
    r_{t}^{p} = \text{exp}(-\lambda^{p} *e_{t}^{p}), \quad e_{t}^{p} = \text{MSE}(\boldsymbol{s}_{t}^{p},\hat{\boldsymbol{s}}_{t}^{p}),
    \label{eq: rp}
\end{aligned}
\end{equation}
\begin{equation}
\begin{aligned}
    r_{t}^{r} = \text{exp}(-\lambda^{r} *e_{t}^{r}), \quad e_{t}^{r} = \text{MSE}(\boldsymbol{s}_{t}^{r},\hat{\boldsymbol{s}}_{t}^{r}),
    \label{eq: rr}
\end{aligned}
\end{equation}
\begin{equation}
\begin{aligned}
    r_{t}^{pv} = \text{exp}(-\lambda^{pv} *e_{t}^{pv}), \quad e_{t}^{pv} = \text{MSE}(\boldsymbol{s}_{t}^{pv},\hat{\boldsymbol{s}}_{t}^{pv}),
    \label{eq: rpv}
\end{aligned}
\end{equation}
\begin{equation}
\begin{aligned}
    r_{t}^{rv} = \text{exp}(-\lambda^{rv} *e_{t}^{rv}), \quad e_{t}^{rv} = \text{MSE}(\boldsymbol{s}_{t}^{rv},\hat{\boldsymbol{s}}_{t}^{rv}),
    \label{eq: rrv}
\end{aligned}
\end{equation}
$\hat{\boldsymbol{s}}_{t}^{p}$, $\hat{\boldsymbol{s}}_{t}^{r}$, $\hat{\boldsymbol{s}}_{t}^{pv}$, $\hat{\boldsymbol{s}}_{t}^{rv}$ are the components of the reference HOI state: the humanoid position, rotation, velocity, and rotation velocity. $\boldsymbol{s}_{t}^{p}$, $\boldsymbol{s}_{t}^{r}$, $\boldsymbol{s}_{t}^{pv}$, $\boldsymbol{s}_{t}^{rv}$ are calculated from simulation and is aligned with the reference HOI state in format. $\lambda^{p}$, $\lambda^{r}$, $\lambda^{pv}$, $\lambda^{rv}$ are hyperparameters that conditions the sensitivity.

\noindent{\textbf{Object Motion Reward.}} Similar to the human motion reward, the object motion reward is:
\begin{equation}
\begin{aligned}
    r_{t}^{o} = r_{t}^{op}*r_{t}^{or}*r_{t}^{opv}*r_{t}^{orv},
    \label{eq: ro}
\end{aligned}
\end{equation}
where $r_{t}^{op}$, $r_{t}^{or}$, $r_{t}^{opv}$, $r_{t}^{orv}$ are the object position reward, rotation reward, position velocity reward, and rotation velocity reward, respectively. The calculation of these rewards is similar to the Eq.~\ref{eq: rp}, with $\lambda^{op}$, $\lambda^{or}$, $\lambda^{opv}$, $\lambda^{orv}$ denotes the hyperparameters that condition the reward sensitivity.

\noindent{\textbf{Interaction Graph Reward.}} The reward for the IG is:
\begin{equation}
\begin{aligned}
    r_{t}^{ig} = \text{exp}(-\lambda^{ig} *e_{t}^{ig}), \quad e_{t}^{ig} = \text{MSE}(\boldsymbol{s}_{t}^{ig},\hat{\boldsymbol{s}}_{t}^{ig}),
    \label{eq: rig}
\end{aligned}
\end{equation}
where $\hat{\boldsymbol{s}}_{t}^{ig}$ is the reference and $\boldsymbol{s}_{t}^{ig}$ is calculated from current simulation. Note that the body motion reward $r_{t}^{b}$, object motion reward $r_{t}^{o}$, and IG reward $r_{t}^{ig}$ are all measurements of kinematic properties. We call them kinematic rewards.

\noindent{\textbf{Contact Graph Reward.}} \label{sec:cgr}
Though kinematic rewards can handle simple interactions, they usually do not handle dynamic scenarios and are not robust to noisy data, \eg, during the training of grasp tasks, the contact between the humanoid and the object will, in most cases, cause the object to move away from the desired trajectory and the expected return becomes smaller. In this case, the policy may learn not to touch the object and falls into a local optimal, as demonstrated in Fig.~\ref{fig: ablation on cgr} (8). We observe that these issues are highly related to unwanted contacts. To encourage accurate contact with the object, we design the contact graph (Sec.~\ref{sec: cg}) and a corresponding contact graph reward (CGR) to guide the humanoid to learn correct contact. The CG error is defined as
\begin{equation}
\begin{aligned}
   \boldsymbol{e}_{t}^{cg} = |\boldsymbol{s}_{t}^{cg}-\hat{\boldsymbol{s}}_{t}^{cg}|,
    \label{eq: ecg}
\end{aligned}
\end{equation}
where $|\cdot|$ represent element wise absolute value. The CGR is measured by CG error, with independent weights on different edges:
\begin{equation}
\begin{aligned}
    r_{t}^{cg} = \text{exp}(-\sum_{j=1}^{J}\boldsymbol{\lambda}^{cg}[j] *\boldsymbol{e}_{t}^{cg}[j]), 
    \label{eq: rcg}
\end{aligned}
\end{equation}
where $J=k(k-1)/2$ is the CG edge numbers; $\boldsymbol{e}_{t}^{cg}[j]$ is the $j$th element of $\boldsymbol{e}_{t}^{cg}\in\{0,1\}^{J}$, a binary label representing a CG edge error; $\boldsymbol{\lambda}^{cg}[j]$ is the $j$th element of $\boldsymbol{\lambda}^{cg}\in\mathbb{R}^{J}$, a hyperparameter controls the sensitivity of a CG edge.
We multiply the CGR with the previous kinematic rewards. With the guidance of CGR, the humanoid learns the correct contact and avoids the local optimal of kinematic rewards. 
The ablation study in Fig.~\ref{fig: ablation on cgr} and Tab.~\ref{tab:evaluation} justifies our design.

\subsection{State}
\label{sec:state}
The state $\boldsymbol{s}_{t}$ is the input of the policy, consisting of two parts: the simulated HOI state $\boldsymbol{g}_{t}$ and the reference HOI state $\hat{\boldsymbol{h}}_{t+1}$ (Sec.~\ref{sec:hoi representation}): 
\begin{equation}
\begin{aligned}
    \boldsymbol{s}_{t} = \{\boldsymbol{g}_{t}, \hat{\boldsymbol{h}}_{t+1} \},
    \label{eq: st}
    \vspace{-0.4cm}
\end{aligned}
\end{equation}

where the reference HOI state $\hat{\boldsymbol{h}}_{t+1}$ is the target that we expect the humanoid and objects to reach in the next time step. 
The simulated HOI state represents the current humanoid and object state under simulation and is necessary for the policy to sense the current environment. Similar to previous works \cite{DeepMimic}, we transform all coordinates into the root local coordinate of the humanoid, simplifying data distribution and making it more conducive to learning. For the humanoid, we observe its global root height, local body position, rotation, position velocity, and rotation velocity. These representations form the humanoid (subject) observation $\boldsymbol{o}^{sbj}_{t}$. In addition, we detect net contact forces $\boldsymbol{o}^{f}_{t}$ for the contact bodies (e.g., fingers), which helps to identify contact and accelerate training. For objects, we observe their local position, rotation, velocity, and rotation velocity, which form the object observation $\boldsymbol{o}^{obj}_{t}$. The simulated HOI state can be formulated as:
\begin{equation}
\begin{aligned}
    \boldsymbol{g}_{t} = \{\boldsymbol{o}^{sbj}_{t},\boldsymbol{o}^{f}_{t},\boldsymbol{o}^{obj}_{t}\}.
    \label{eq: gt}
\end{aligned}
\end{equation}

\subsection{Policy and Action}
\label{sec: policy and action}
We follow the actor-critic framework widely used by prior arts \cite{amp,ase}. The policy output is modeled as a Gaussian distribution of dimensions $51\times 3$ with constant variance, and the mean is modeled by a two-layer MLP of [1024, 512] units and ReLU activations. The action $\boldsymbol{a}_{t}\in\mathbb{R}^{51\times 3}$ sampled from the policy is the target joint rotations for the PD controller. The PD controller adjusts and outputs the joint torques to reach the target rotations.

\subsection{Simulation Setting}
\label{sec:simulation}
The simulation and the PD controller run at 60 Hz. The policy is sampled at 30Hz. The humanoid and objects need initialization when the simulation starts. We use fixed initialization by default; that is, we extract the rotations and root positions from the first reference frame to initialize the humanoids and objects. We do not use random initialization since the HOI data may have severe collisions that eject the object. We use early termination depending on the time and kinematic state errors. See the Appendix for details.

\section{The BallPlay Dataset}
\label{sec: ballplay}
\begin{figure}[h]
  \centering
  \includegraphics[width=1\linewidth]{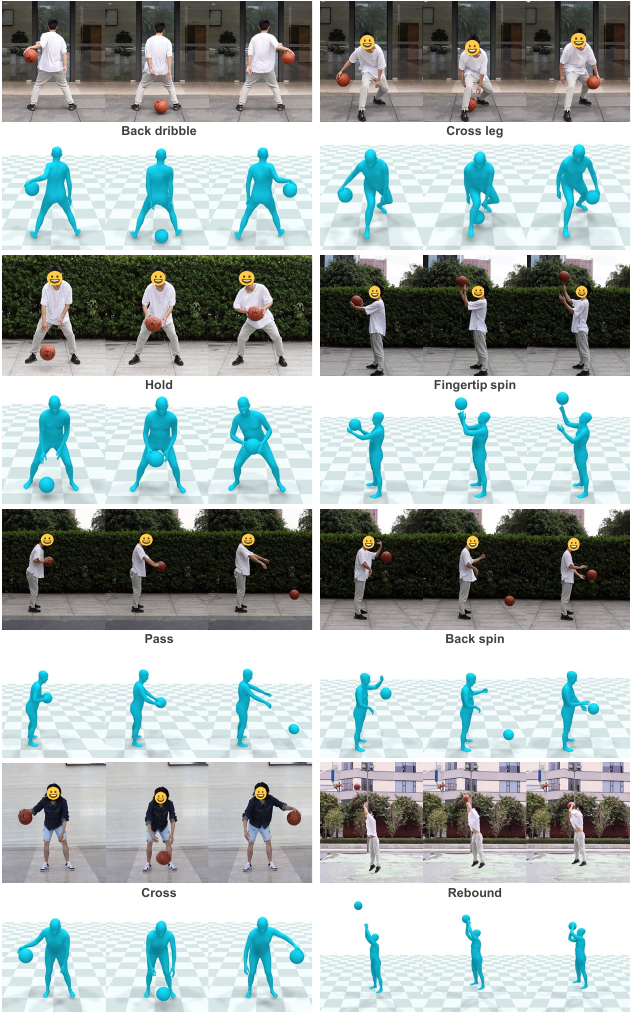}
  \caption{\textbf{The \emph{BallPlay} dataset.} We show the eight HOI demonstrations of high-dynamic basketball skills. For each skill, the \textbf{upper} rows show the real-life videos, and the \textbf{lower} rows give the estimated whole-body SMPL-X human model and object mesh.}
  \vspace{-0.5cm}
\label{fig: ballplay} 
\end{figure}

To compensate for the lack of dynamic HOI scenarios, we introduce the \emph{BallPlay} dataset containing \textbf{eight} whole-body basketball skills, including \emph{back dribble, cross leg, hold, fingertip spin, pass, backspin, cross, and rebound}, as shown in Fig.~\ref{fig: ballplay}. Instead of using MoCap devices that are costly and hard to scale up, we apply a monocular annotation solution to estimate the high-quality human SMPL-X parameters and object translations from RGB videos. However, annotating these videos with high-speed and dynamic movements and complex interactions in the 3D camera coordinate is quite challenging. 
Inspired by the whole-body annotation pipeline of Motion-X~\cite{lin2023motionx}, our automatic annotation additionally introduces depth estimation~\cite{bhat2023zoedepth}, semantic segmentation~\cite{Grounded-SAM}, and CAD model selection to obtain high-quality whole-body human motions and object motions.   
Besides, we manually annotate three key contact frames for each video to help the object depth estimation. 
The dataset contains videos, estimated HOI sequences, contact labels, and physically rectified HOI sequences using \ModelName. Annotation details can be found in the Appendix. 

\begin{figure*}[t]
  \centering
  \vspace{-0.5cm}
  \includegraphics[width=1\linewidth]{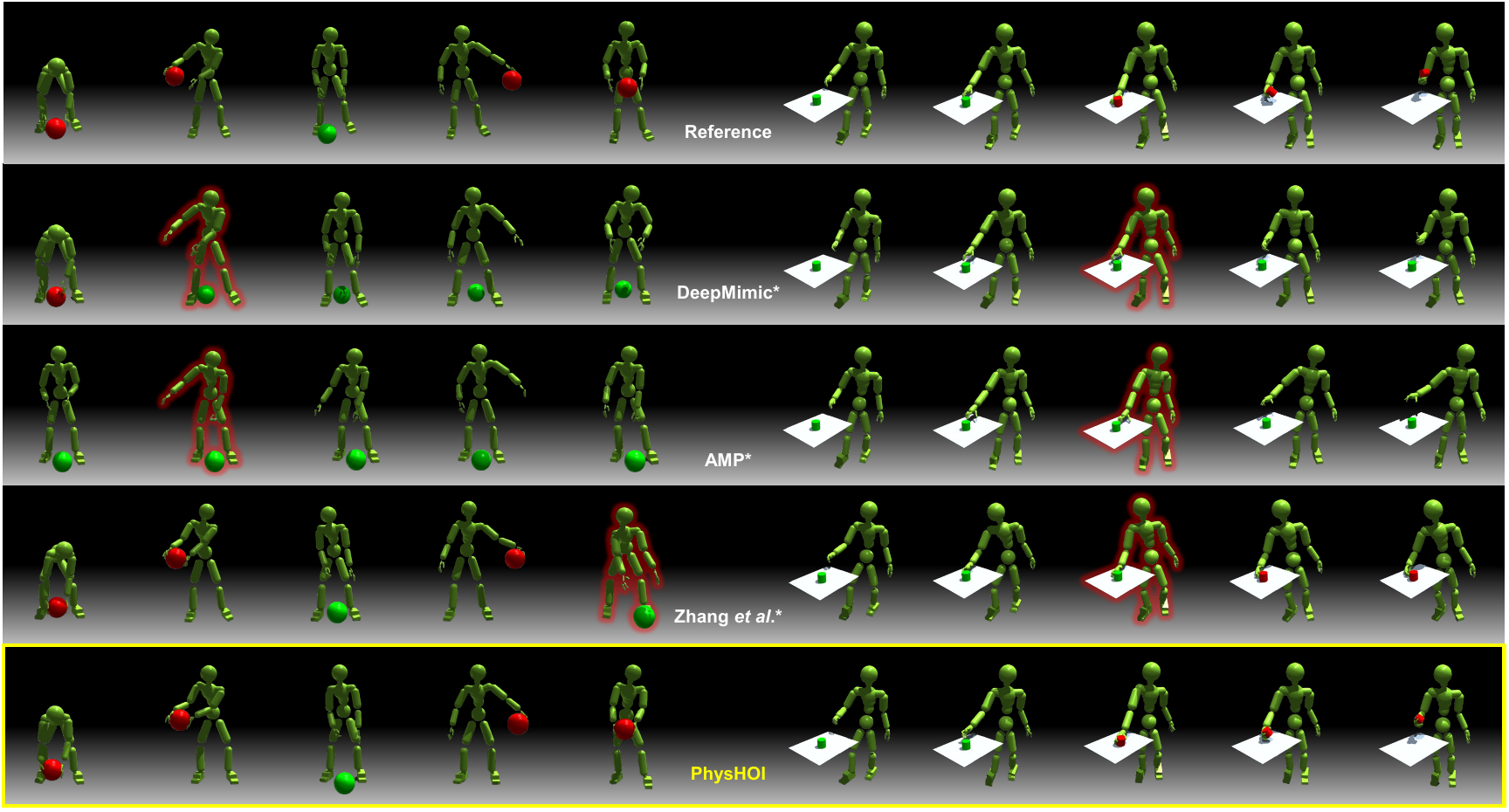}
  \vspace{-0.5cm}
  \caption{\textbf{Qualitative results on HOI imitation.} *means re-implemented methods. Previous methods that use kinematic-only rewards fail to reproduce the interaction accurately, e.g., the ball falls or the grasp fails. Guided by the contact graph, our method yields successful HOI imitation. We mark the object red when it has contact with the humanoid. We outline in \textcolor{red}{red} the frame where the failure begins. }
  \vspace{-0.4cm}
\label{fig: compare} 
\end{figure*}

\section{Experiment}
\subsection{Evaluation on HOI Imitation}

\noindent{\textbf{Datasets.}}
We evaluate \ModelName on two types of HOI datasets: GRAB \cite{GRAB2020} and our proposed BallPlay (Sec.~\ref{sec: ballplay}).
We chose 5 cases from the GRAB S8 subset, including \emph{grasping cube, cylinder, flashlight, flute, and bottle}.

\noindent{\textbf{Metrics.}}
We report three types of metrics: 1) the success rate (Succ) of HOI imitation, where the success is defined per frame, deeming imitation successful when the object position and body position errors are both under the thresholds and the contact graph edge value is correct. The Succ is calculated by averaging the success values of all frames. The object threshold is defined as 0.2$m$. The body threshold is defined as 0.1$m$.
The Succ reflects whether the humanoid can accurately imitate the reference body motion to interact with the object correctly. 2) the mean per-joint position error (MPJPE) of the humanoid ($E_\text{b-mpjpe}$) and object ($E_\text{o-mpjpe}$) to evaluate the positional tracking performance (in $mm$) following ~\cite{luo2023perpetual}. 
3) the contact accuracy $E_\text{cg}$, ranging from 0 to 1, defined as $\frac{1}{N}\sum_{t=1}^{N}\text{MSE}(\boldsymbol{s}_{t}^{cg},\hat{\boldsymbol{s}}_{t}^{cg})$ where N is the total frames of the reference HOI data.

\begin{figure*}[t]
  \centering
  \vspace{-0.3cm}
  \includegraphics[width=1.\linewidth]{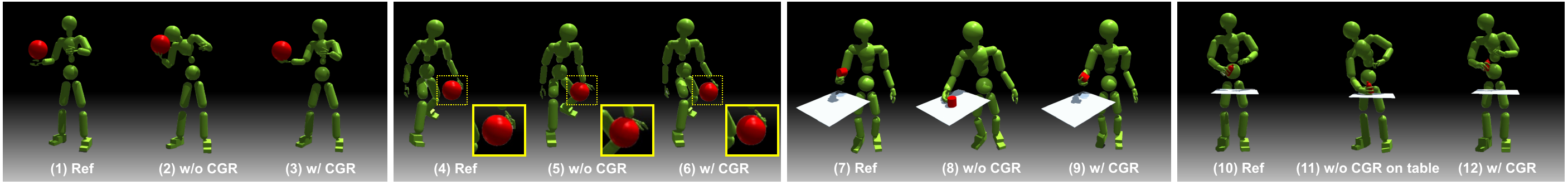}
  \vspace{-0.5cm}
  \caption{\textbf{Ablation on Contact Graph Reward (CGR).} Without CGR, the humanoid is easy to fall into local optimal, e.g., (\textbf{2}) using the head to help control the ball, (\textbf{5}) using the wrist to touch the ball, (\textbf{8}) being afraid of catching objects, (\textbf{11}) supporting the table to keep balance. In comparison, the use of CGR effectively guides the humanoid toward correct interaction, as shown in (\textbf{3}), (\textbf{6}), (\textbf{9}), and (\textbf{12}).}
  \vspace{-0.1cm}
\label{fig: ablation on cgr} 
\end{figure*}

\begin{table*}[t]
\centering
\resizebox{\linewidth}{!}{%
\begin{tabular}{lr|rrrr|rrrr|rrrr}
\toprule
\multicolumn{2}{c}{} & \multicolumn{4}{c}{DeepMimic* } & \multicolumn{4}{c}{Zhang \etal*} & \multicolumn{4}{c}{\textbf{\ModelName (Ours)}} 
\\ 
\midrule
Dataset  &  & $\text{Succ} \uparrow$ & $E_\text{b-mpjpe}  \downarrow$ &  $E_\text{o-mpjpe} \downarrow $ &  $E_\text{cg} \downarrow$   & $\text{Succ} \uparrow$ & $E_\text{b-mpjpe} \downarrow$ & $E_\text{o-mpjpe} \downarrow $ &  $E_\text{cg} \downarrow$   & $\text{Succ} \uparrow$ & $E_\text{b-mpjpe} \downarrow$ &  $E_\text{o-mpjpe} \downarrow $ &  $E_\text{cg} \downarrow$       \\ \midrule
\emph{GRAB} &  &  {27.0\%} & \textbf{44.7} & {180.2}  & {0.7240} &  {38.6\%} & {91.0} & {180.1 }  & {0.3370} & \textbf{95.4\%} & {71.1} & \textbf{78.0} & \textbf{0.0260} \\
\emph{BallPlay} &  &  {7.5\%} & \textbf{40.1} & {1662.5}  & {0.3063} &  {13.6\%} & {88.5} & {155.3}  & {0.4124} &  \textbf{82.4\%} & {56.8} & \textbf{82.9} & \textbf{0.0877} \\
\midrule
\end{tabular}}
\vspace{-0.2cm}
    \caption{\textbf{Quantitative results on HOI imitation.} Our method yields superior object control and success rate on HOI imitation. *means re-implemented methods. 
    We repeat all sequences 10 times, report the average values, and \textbf{bold} the best score.}
\label{tab:evaluation}
\vspace{-0.1cm}
\end{table*}

\begin{table*}[t!]
\centering
\resizebox{\linewidth}{!}{%
\begin{tabular}{lr|rrrr|rrrr|rrrr}
\toprule
\multicolumn{2}{c}{} & \multicolumn{4}{c}{$r_{t}^{b}*r_{t}^{o}$ } & \multicolumn{4}{c}{$r_{t}^{b}*r_{t}^{o}*r_{t}^{ig} $} & \multicolumn{4}{c}{$r_{t}^{b}*r_{t}^{o}*r_{t}^{ig}*r_{t}^{cg}$} 
\\ 
\midrule
HOI Data  &  & $\text{Succ} \uparrow$ & $E_\text{b-mpjpe}  \downarrow$ &  $E_\text{o-mpjpe} \downarrow $ &  $E_\text{cg} \downarrow$   & $\text{Succ} \uparrow$ & $E_\text{b-mpjpe} \downarrow$ & $E_\text{o-mpjpe} \downarrow $ &  $E_\text{cg} \downarrow$   & $\text{Succ} \uparrow$ & $E_\text{b-mpjpe} \downarrow$ &  $E_\text{o-mpjpe} \downarrow $ &  $E_\text{cg} \downarrow$       \\ \midrule
\emph{Toss} &  &  {20.7$\%$} & {79.9} & \textbf{31.9}  & {0.3837} &  {7.7$\%$} & {86.9} & {154.3}  & {0.4396} & \textbf{84.8\%} & \textbf{45.7} & {77.9} & \textbf{0.0767} \\
\emph{Cross leg} &  &  {6.7$\%$} & {292.4} & {246.2}  & {0.5189}  & {8.2$\%$} & \textbf{42.0} & \textbf{51.6}  & {0.5394} & \textbf{88.2\%} & {54.1} & {92.3} & \textbf{0.0843} \\
\emph{Backspin} &  &  {0.2$\%$} & {1247.8} & {7105.6}  & {0.5460} &  {21.7$\%$} & \textbf{50.4} & {74.9 }  & {0.3908} &  \textbf{70.3\%} & {60.2} & \textbf{65.2} & \textbf{0.1302} \\
\emph{Pass} &  &  {4.8$\%$} & {616.5} & {1047.7}  & {0.5118} &  {16.2$\%$} & {203.8} & {418.0 }  & {0.2909}  & \textbf{71.2\%} & \textbf{72.5} & \textbf{110.6} & \textbf{0.1135} \\
\midrule
\end{tabular}}
\vspace{-0.2cm}
    \caption{\textbf{Ablation on rewards.} The interaction graph reward $r_{t}^{ig}$ can refine the interaction and motion imitation quality. The contact graph reward $r_{t}^{cg}$ significantly improves the overall performance, especially the success rate, on HOI imitation. }
\label{tab:ablation}
\vspace{-0.2cm}
\end{table*}

\noindent{\textbf{Baseline Methods.}}
To the best of our knowledge, our proposed \ModelName is the first whole-body HOI imitation method. For fair comparisons, we make proper adaptations for previous human motion imitation methods to experiment on HOI imitation tasks: (1) DeepMimic \cite{DeepMimic} is a motion imitation method using additive motion rewards; we add the object motion reward to it; (2) AMP \cite{amp} is an unaligned motion imitation method, we modify its reference state as our proposed contact-aware HOI representation from Sec.\ref{sec:hoi representation}; (3) We implement a simplified version of Zhang's method \cite{ig2023} via our interaction graph (Sec.~\ref{sec:hoi representation}) since the code is not available yet.

\noindent{\textbf{Implementation Details.}}
We use Isaac Gym \cite{makoviychuk2021isaac} as the physics simulation platform. All experiments are trained on a single Nvidia A100 GPU, with 2048 parallel environments and fixed simulation initialization. We train 5000 epochs for all experiments on the GRAB dataset, and 15000 epochs for all experiments on the BallPlay dataset. Each epoch contains 10 frames of sequential simulation. We use the aggregated contact graph (CG) for experiments. For GRAB, the CG contains 3 nodes: the table, the object, and the aggregated whole body. For BallPlay, the CG contains 3 nodes: the object, aggregated hands, and the aggregated rest body parts. 
We do not consider the basketball rotation since it is not provided, \ie, we set $\lambda^{or}$ and $\lambda^{orv}$ as zero for experiments on BallPlay. 
The setting of hyperparameters can be found in the Appendix.

\noindent{\textbf{Qualitative Results.}} From Fig.~\ref{fig: compare}, DeepMimic \cite{DeepMimic} and AMP \cite{amp} achieve reasonable humanoid motion but fail to control the object. 
Zhang \etal \cite{ig2023} is effective on some interactions but is also prone to local optima due to the lack of contact guidance. We highlight the failure beginning frames in red outlines. 
In contrast, the proposed \ModelName introduces the contact graph that makes it able to handle various complex interactions.

\noindent{\textbf{Quantitative Results.}}
Tab.~\ref{tab:evaluation} shows the average metrics on the GRAB subset and the BallPlay dataset. DeepMimic \cite{DeepMimic} achieves the best score in body motion metrics, \ie, $E_\text{b-mpjpe}$, since it only learns human motions. It fails to control the object and results in low interaction success rates. Zhang \etal \cite{ig2023} achieves better interaction than DeepMimic, but still yields low success rates because only learning kinematic information. Interestingly, their body imitation errors increase severely, and the contact accuracy worsens for dynamic HOI. 
In contrast, \ModelName with CG provides effective contact guidance on HOI learning and yields superior success rates.

\vspace{-0.1cm}
\subsection{Ablation on Contact Graph Reward}
Tab.~\ref{tab:ablation} and Fig.~\ref{fig: ablation on cgr} study the effectiveness of CGR. Overall, the use of CGR significantly improves success rates. If it only applies kinematic reward, the humanoid is prone to fall into local optimal.  For instance, the kinematic rewards result in decent kinematic metrics on the \emph{Cross leg} data, but it uses false body parts to control the ball, resulting in inferior HOI imitation in Fig.~\ref{fig: ablation on cgr} (5). Similarly, in Fig.~\ref{fig: ablation on cgr} (2), on the \emph{Toss} data, the humanoid wrongly learns to use its head to control the ball. On the contrary, by applying the contact graph reward in Fig.~\ref{fig: ablation on cgr} (3) and (6), the humanoid learns correct contact and avoids incorrect collisions. Another interesting phenomenon is that the HOI generated by our methods is more accurate than the reference. From Fig.~\ref{fig: ablation on cgr} (1), there is a slight floating between the hand and the ball in the reference, while our result fits perfectly.

Considering contact with multiple objects is also necessary. In the grasp tasks, the humanoid sometimes learns to use its hands to support the table to maintain balance if the table is not considered in the contact graph, as shown in Fig.~\ref{fig: ablation on cgr} (11). By involving the table in the contact graph, this local optimal can be well addressed (Fig.~\ref{fig: ablation on cgr} (12)). In summary, comprehensive studies have validated the essence and effectiveness of the proposed CGR.
\vspace{-0.1cm}
\section{Conclusion}
We present a framework for learning Human-Object Interaction (HOI) skills from HOI demonstrations. Our method is able to imitate a diverse set of highly dynamic basketball skills. We involve the contact graph (CG) to guide the humanoid toward precise object control, which is proven to be critical for HOI imitation and is effective even when the reference HOI data is highly biased. The proposed task-agnostic HOI imitation reward is effective on diverse HOI types without the need to design task-specific rewards. 
We also introduce an HOI dataset called BallPlay is provided to support research on dynamic HOI. We believe this work opens up many exciting directions for future exploration toward general HOI learning. See the Appendix for more discussions about limitations and future work. 

\appendix
{
    \small
    \bibliographystyle{ieeenat_fullname}
    \bibliography{main}
}

\clearpage
\setcounter{page}{1}
\maketitlesupplementary

In the following document, we provide additional information to supplement the main paper. Video demonstrations can be found on our project page. We will release the BallPlay dataset and open source the code.

\etocdepthtag.toc{mtappendix}
\etocsettagdepth{mtchapter}{none}
\etocsettagdepth{mtappendix}{subsection}
{
  \hypersetup{
    linkcolor = black
  }
  \tableofcontents
}

\section{Technical Details}
\label{app: technical details}

\subsection{HOI Data Preprocessing} 

Since the shapes of humans and objects in HOI data are represented by mesh and cannot be directly used for the simulation environment, we build the simulation models of robots and objects to match their meshes and make necessary calibrations for the HOI data.

\noindent{\textbf{{Raw HOI Data Format.}} The raw HOI data consists of frames at 30 fps. Each frame contains the human joint rotation, human root rotation, human root position, object position, and object rotation. We use the SMPL-X model \cite{SMPL-X} to parameterize the whole-body shape as $\mathcal{
\beta}\in \mathbb{R}^{10}$ and whole-body pose as $\mathcal{
\theta}\in \mathbb{R}^{51\times3}$. The object is represented as mesh. 

\begin{figure}[h]
  \centering
  \includegraphics[width=1\linewidth]{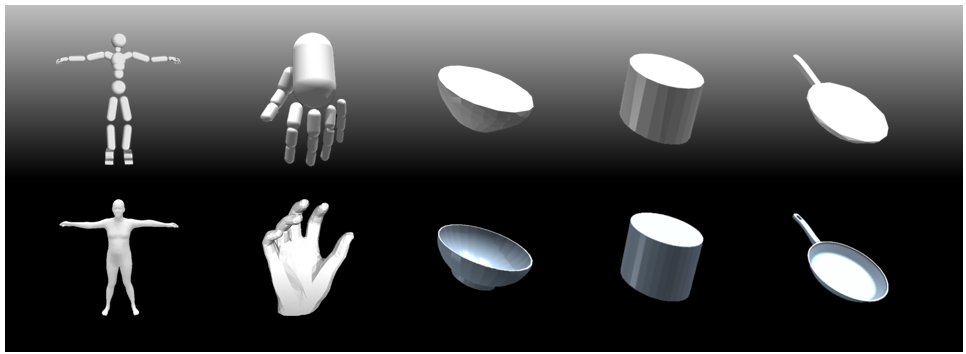}
  \caption{The original human and object shape (\textbf{bottom}) and their approximated simulation model (\textbf{up}).}
  \vspace{-0.2cm}
\label{fig:models} 
\end{figure}

\noindent{\textbf{Simulation Models for Human and Object.}}
Given an SMPL-X shape parameter $\mathcal{
\beta}$, we generate a corresponding whole-body humanoid robot following UHC \cite{Luo2021DynamicsRegulatedKP, Luo2022EmbodiedSH}. To drive the poses, we use SMPL-X pose parameter $\mathcal{
\theta}$. Accordingly, the robot has a total of 51$\times$3 DoF actuators where 21$\times$3 DoF for the body and 30$\times$3 DoF for the hands. To simplify collision calculation, bodies are simplified as simple geometries like capsules or boxes. For general objects, we use convex decomposition to approximate the object mesh as convex hulls, which is necessary for collision detection. For the basketball, we simply use a sphere as an approximation. Fig.~\ref{fig:models} shows the original human and object shape and the approximated simulation model.

\noindent{\textbf{HOI Data Calibration.}} 
We transfer the original HOI data to the simulation environment and perform coordinate calibration to obtain preliminary data samples. To accurately model interaction information, we extract and save the body positions and the binary contact labels per frame, which can be easily acquired by reading the simulator's API after loading the calibrated HOI data into the simulation. This data calibration process is effective in resolving hand-object penetration in the data, because the simulator will automatically adjust the body position to avoid collision.

\subsection{Details on Simulation Setting}

\noindent{\textbf{Simulation Initialization.}} 
We use first-frame initialization by default; that is, we extract the rotations and root positions from the first reference state to initialize the robot and objects. 

\noindent{\textbf{Early Termination.}}
Since our imitation learning is strictly aligned with the reference frame, the maximum duration of the simulation is the length of the reference HOI sequence. In general, during training, the simulation is reset only when the maximum time is reached. However, during the training process, the robot often fails before reaching the maximum time, such as falling down or the object significantly deviating from the reference trajectory. In these cases, we reset the simulation to improve the simulation sample efficiency. Specifically, in three cases we reset the simulation: (1) the maximum time is reached; (2) the object deviates far from the reference trajectory; (3) the robot positions deviate from the reference. The threshold of position error is set to 0.5m.

\subsection{Experiment Details}
\label{sec: exp details}

\noindent{\textbf{Hyperparameters.}}
We use the aggregated contact graph (CG) for experiments. For GRAB, the CG contains 3 nodes: the table, the object, and the aggregated whole body. For BallPlay, the CG contains 3 nodes: the object, the aggregated hands, and the aggregated rest body parts. We provide the setting of reward weights in Tab.~\ref{tab:hyperparameters}. For \emph{BallPlay}, $\boldsymbol{\lambda}^{cg}[0]$ corresponding to the CG edge connecting the hands and the ball, $\boldsymbol{\lambda}^{cg}[1]$ corresponding to the CG edge connecting the hands and the rest body, $\boldsymbol{\lambda}^{cg}[2]$ corresponding to the CG edge connecting the rest body and the ball. For \emph{GRAB}, $\boldsymbol{\lambda}^{cg}[0]$ corresponding to the CG edge connecting the body and the object, $\boldsymbol{\lambda}^{cg}[1]$ corresponding to the CG edge connecting the body and the table, $\boldsymbol{\lambda}^{cg}[2]$ corresponding to the CG edge connecting the table and the object.

\begin{table}[t!]
\centering
\resizebox{\linewidth}{!}{%
\begin{tabular}{l|llll|llll|l|llll}
\toprule
\multicolumn{1}{c}{} & \multicolumn{4}{c}{\text{body} } & \multicolumn{4}{c}{\text{object}} & \multicolumn{1}{c}{\text{IG}} & \multicolumn{3}{c}{\text{CG}} 
\\ 
\midrule
Dataset  &  $\lambda^{p}$ & $\lambda^{r}$ &  $\lambda^{pv}$ &  $\lambda^{rv}$   & $\lambda^{op}$ & $\lambda^{or}$ & $\lambda^{opv}$ &  $\lambda^{orv}$   & $\lambda^{ig}$ & $\boldsymbol{\lambda}^{cg}[0]$ &  $\boldsymbol{\lambda}^{cg}[1]$ &  $\boldsymbol{\lambda}^{cg}[2]$       \\ \midrule
\emph{BallPlay}   &  {50} & {20} & {0.01}  & {0.01} & {1} & {0} & {0.01} & {0}  & {20} & {5} & {5} & {5} \\
\emph{GRAB}  &  {50} & {20} & {0.01}  & {0.01} & {1} & {0.1} & {0.01} & {0.01}  & {20} & {50} & {5} & {5} \\
\midrule
\end{tabular}
}
    \caption{\textbf{Reward weights for PhysHOI.} Note that for the \emph{fingertip spin} case, we set $\boldsymbol{\lambda}^{cg}[0]$=0.01 to weak restrictions on contact between hands and the ball, as explained in Fig.~\ref{fig: failurecase}.}
\label{tab:hyperparameters}
\end{table}

\begin{table}[t!]
\centering
\resizebox{\linewidth}{!}{%
\begin{tabular}{l|c|llll|llll|l|l}
\toprule
\multicolumn{1}{c}{} & \multicolumn{1}{c}{}& \multicolumn{4}{c}{\text{body} } & \multicolumn{4}{c}{\text{object}} & \multicolumn{1}{c}{\text{IG}} & \multicolumn{1}{c}{\text{CG}} 
\\ 
\midrule
Methods  & + \text{or} $\times$ &  $r^{p}$ & $r^{r}$ &  $r^{pv}$ &  $r^{rv}$   & $r^{op}$ & $r^{or}$ & $r^{opv}$ &  $r^{orv}$   & $r^{ig}$ & $r^{cg}$       \\ \midrule
\text{DeepMimic}  & + &  \greencheck & \greencheck & \greencheck  & \greencheck & \greencheck & \greencheck & \greencheck & \greencheck  & \redcross & \redcross \\
Zhang \etal &  $\times$  &  \greencheck & \greencheck & \greencheck  & \greencheck & \greencheck & \greencheck & \greencheck & \greencheck  & \greencheck & \redcross \\
\midrule
\end{tabular}
}
    \caption{Reward design for re-implemented methods.}
\label{tab:re-implemented methods}
\end{table}

\noindent{\textbf{Details on Re-Implemented Methods.}}
For fair comparisons, we re-implement related state-of-the-art methods: DeepMimic \cite{DeepMimic}, AMP \cite{amp}, and Zhang's method \cite{ig2023}. The main difference is the reward design, which is presented in Tab.~\ref{tab:re-implemented methods}. Note that our code is based on the ASE project \cite{ase}, which contains the implementation of AMP. We change the original AMP observation as our HOI representation for HOI imitation.

\noindent{\textbf{Contact Detection.}}
Since Isaac Gym does not yet provide contact detection APIs for the GPU pipeline, we use force detections as approximations for contact detections.  For example, if the net contact force of the ball on either of the x and y axes is above a threshold and the net contact force of the hands-excluded body parts is under a threshold, we deem that the ball has contact with the hands.

\section{The BallPlay Dataset}
\label{sec: ballplay details}

\subsection{HOI Data Annotation}

To construct the BallPlay dataset, we propose a semi-automatic annotation pipeline to estimate 3D human-object interaction motion from monocular RGB videos. As depicted in Fig.~\ref{fig:annotation}, our annotation pipeline mainly involved three stages: whole-body human motion annotation, object annotation, and contact-based manual correction.

\noindent{\textbf{Whole-Body Human Motion Annotation.}} Initially, we adopt an optimization-based approach, as introduced in Motion-X~\cite{lin2023motionx}, to annotate high-quality 3D whole-body human motion from RGB videos. This optimization process also includes the estimation of camera parameters.

\noindent{\textbf{Object Annotation.}} Subsequently, we employ Grounded-SAM \cite{Grounded-SAM} to annotate the object category and the segmentation mask, along with utilizing ZoeDepth~\cite{bhat2023zoedepth} for depth estimation. We then project the object mask into a point cloud in the world coordinate system based on the depth map and camera parameters. The central point of the resulting point cloud serves as the initialized object center. Meanwhile, based on the object category, we retrieve the object mesh from a CAD model library. The meshes of the CAD model library are collected from ShapeNet~\cite{chang2015shapenet} and some geometries built by ourselves with Trimesh~\cite{trimesh} tool. 

\noindent{\textbf{Contact-based Manual Correction.}} Due to the inherent depth ambiguity of monocular video, the object center obtained from the previous steps can be noisy. To address this issue, we propose a contact-based manual correction procedure. Specifically, we manually annotate the human-object contact labels of three keyframes within each video, and then compute the average depth of the contact body parts to update the object center position. This manual correction can greatly mitigate depth ambiguity and enhance overall data quality. Ultimately, our refined object mesh and human mesh constitute the final human-object interaction dataset.

\begin{figure}[h]
  \centering
  \includegraphics[width=1\linewidth]{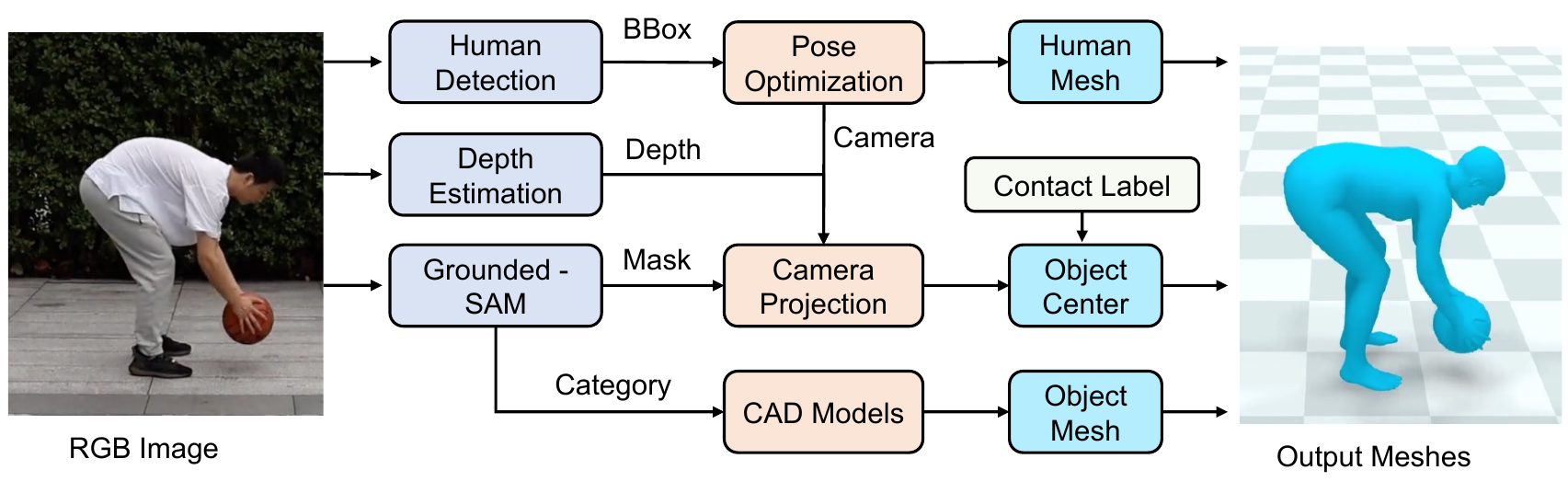}
  \caption{Annotation pipeline of \emph{BallPlay}.}
\label{fig:annotation} 
\end{figure}

\begin{figure*}[h]
  \centering
  \vspace{-0.5cm}
  \includegraphics[width=1.\linewidth]{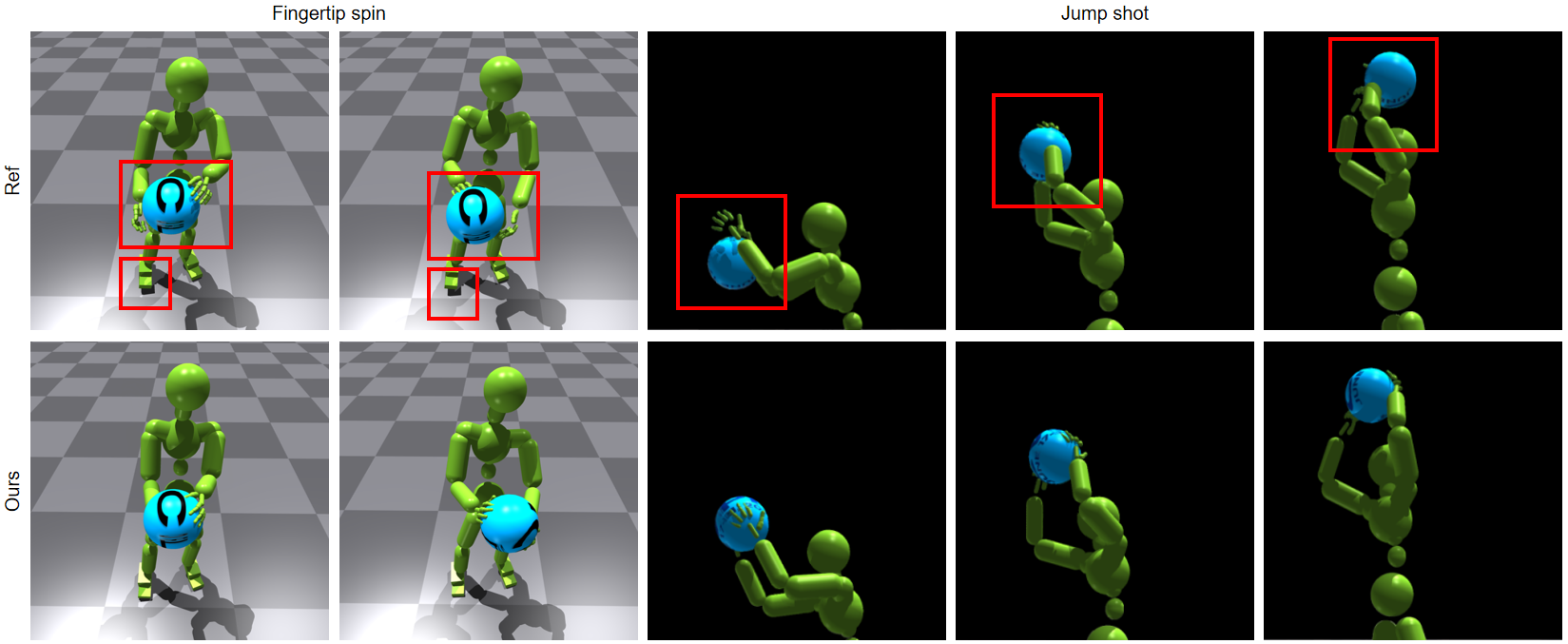}
  \caption{\textbf{PhysHOI can rectify the error of reference HOI data}. We can see that the reference HOI data suffers inaccuracies more or less, while the result generated by PhysHOI can be much better than the reference.}
\label{fig: rec} 
\end{figure*}

\noindent{\textbf{Contact Graph Annotation.}}
We focus on basketball skills that involve only hands to interact with the ball. In this scenario, we aggregate all elements as 3 Contact Graph (CG) nodes: the ball, the hands, and the rest of body parts. For all sequences, the rest body parts do not come into contact with the ball and the hands, so we only need to manually label the CG edge between the hands and the ball, which is easy to obtain through observation. 

\subsection{Refine The HOI Data Using PhysHOI}
Though some HOI data in BallPlay may be biased, we may resort to PhysHOI to eliminate errors and yield physically plausible HOI data. As shown in Fig.~\ref{fig: rec}, we can see that the reference HOI data suffer inaccuracies, while the results generated by PhysHOI are much better than the reference, in terms of physical plausibility and smoothness. Specifically, we first train PhysHOI to imitate an HOI sequence. After convergence, we do the inference under physics simulation and record the HOI data as well as the CG through the API of the environment. Physically rectified HOI data is also provided in the BallPlay dataset.

\section{Additional Experiments}

\subsection{Varying Ball Sizes}
During the inference time, we change the radius of the ball to different values and surprisingly find that the humanoids can still perform plausible interaction, even if they are not trained on these ball sizes. Fig.~\ref{fig: ablation_radius} shows the result in three cases. Interestingly, we still get reasonable results without additional training for different ball sizes, which shows the robustness of PhysHOI. Note that the humanoid may fail if the ball size changes too drastically.

\subsection{Ablation on Contact Graph Reward}
In addition to the ablation in the main paper, we provide more visual comparisons in Fig.~\ref{fig: fullcompare} for an intuitive understanding. In the \emph{cross leg} case, the humanoid trained w/o CGR tends to hold the ball between its left hand and leg (we provide another view in red boxes for better observation), which causes unnatural interaction. In the \emph{pass} case, the ball drops without the CGR. In the \emph{fingertip spin} case, the humanoid trained w/o CGR tends to hold the ball with its hands and head. In comparison, applying CGR addresses these problems well. In Fig.~\ref{fig: multienv} we visualize multiple environments to intuitively demonstrate the success rate w/ or w/o CGR. Each environment applies different random seeds. We can see that the ball falling w/o CGR is not accidental, but common. Instead, our method applies CGR and achieves steady success in ball holding.

\section{Related Work on Kinematic-Based Methods}
\subsection{Human Motion Generation}
Kinematic motion generation methods are usually learned from motion capture datasets \cite{kit,amass,humanml3d,lin2023motionx}. Early attempts generate motions from the prefix \cite{Fragkiadaki_Levine_Felsen_Malik_2015, Martinez_Black_Romero_2017} and keyframes \cite{Harvey_Yurick_Nowrouzezahrai_Pal_2020}. \cite{Holden_Saito_Komura_2016} generate motions using an autoencoder, and further apply motion phases to generate consistent motions \cite{Holden_Komura_Saito_2017, starke2020local, Starke_Zhao_Zinno_Komura_2021, Starke_Mason_Komura_2022}. Beyond action-conditioned motion generation \cite{Guo_Zuo_Wang_Zou_Sun_Deng_Gong_Cheng_2020, mvae, Petrovich_Black_Varol_2021}, recent advances make it possible to control motions with text description \cite{humanml3d, mdm, motiondiffuse, zhang2023t2m, chen2023executing,jin2023act,humantomato}, music \cite{aist++, tseng2023edge, dabral2023mofusion}, and speech \cite{yoon2019robots,liang2022seeg,zhu2023taming}, based on recent progress in autoregressive models \cite{gpt2,gpt3} and diffusion models \cite{ho2020denoising,song2021denoising,song2019generative,song2020denoising,song2020score,kawar2022denoising,wang2023zeroshot}.

\subsection{Human-Object Interaction Generation}
There are many research topics surrounding HOI, especially detection~\cite{gkioxari2018detecting,ji2021detecting,wu2022mining,zhou2022human,xie2023visibility,chen2023detecting,zhu2023diagnosing} and reconstruction~\cite{xie2022chore,zhang2020perceiving,wang2022reconstructing,petrov2023object,hou2023compositional,kim2023ncho}. Based on recent development of 3D HOI datasets~\cite{GRAB2020, bhatnagar22behave,jiang2022chairs,huang2022intercap,zhang2023neuraldome,fan2023arctic,starke2019neural, zhang2022couch,hassan_samp_2021}, there emerges a branch of research focuses on generating human motion that interacts with objects \cite{Starke_Zhao_Komura_Zaman_2020, Taheri_Choutas_Black_Tzionas_2022, wu2022saga, ghosh2022imos, lee2023lama, xu2023interdiff, li2023object, tendulkar2023flex, pi2023hierarchical, kapon2023mas}. However, all the methods mentioned learn directly from kinematic datasets without strict modeling of dynamics and contacts, causing artifacts that harm the motion quality, \eg, penetration, shaking, sliding, and floating. Besides, kinematic HOI generation can not be used for robot control yet.

\begin{figure*}[t!]
  \centering
  \vspace{-0.2cm}
  \includegraphics[width=1.\linewidth]{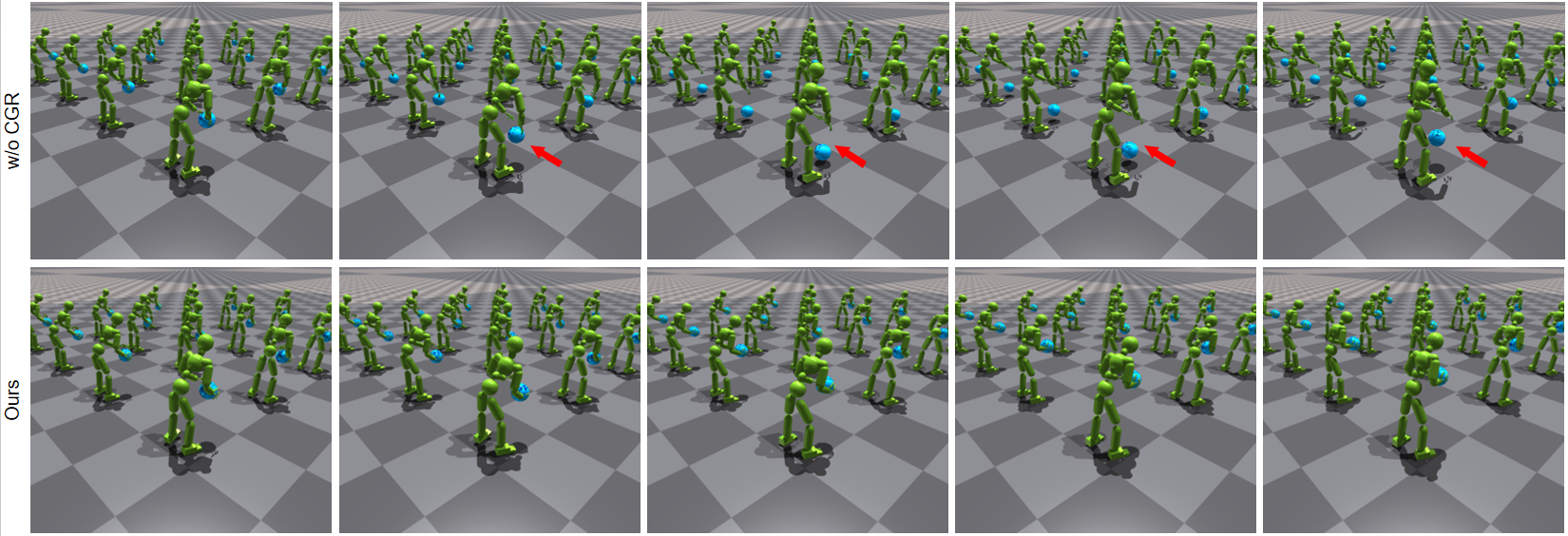}
  \caption{\textbf{Ablation on Contact Graph Reward (CGR)}. We visualize multiple environments to intuitively demonstrate the success rate w/ or w/o CGR. We can see that if w/o CGR, the ball falling is not accidental, but common. Instead, our method applies CGR and achieves steady success in ball holding.}
  \vspace{-0.2cm}
\label{fig: multienv} 
\end{figure*}

\section{Discussions}
\label{sec: discussions}
\subsection{Failure Cases}
The failure cases of our method are mainly due to inaccurate data or incomplete contact graphs. For example, in the \emph{rebound} case in Fig.~\ref{fig: failurecase}, the ball in the reference data is always under the hand, which, combined with the fact that the ball bounces back to the hand very quickly, results in a local optimum: the ball is not firmly grasped during the learning process. This problem can be solved with precise HOI data.

High CGR weight may lead to unpleasant interactions if the CG node is not detailed enough. As shown in the \emph{fingertip spin} case in Fig.~\ref{fig: failurecase}, when we set a high CGR weight on the CG edge between hands and the ball, the trained humanoid tends to use multiple fingers to maintain steady contact. The reasons can be twofold: (1) The CG is not detailed enough, e.g., we simply take two hands as one CG node, but it is necessary to take fingers as separate CG nodes in this case. (2) The frame rate of simulation and reference data is low, which yields frequent bouncing, i.e., less contact. While reducing the corresponding CG edge weights can improve this problem, as shown in the \emph{fingertip spin} case in Fig.~\ref{fig: ablation_radius} and Fig.~\ref{fig: fullcompare}, a more general solution requires richer CGs and higher frame rates, which is also the solution we expect.

\subsection{Limitations}
Though our method is able to mimic diverse dynamic HOI skills, it still faces several limitations, which we list here:
\begin{itemize}
    \item Our method may fail when the reference HOI data suffers severe biases, \eg, the false reference data can result in a ball drop, as shown in the \emph{rebound} case in Fig.~\ref{fig: failurecase}.  
    \item When CG nodes are not detailed enough, some subtle operations can still easily fall into local optimality. For example, fingers should be independent CG nodes when learning complex in-hand manipulations.
    \item Due to the low frame rate of HOI data and simulation frequency, some minor penetrations may appear.
    \item Informing the policy with a single reference object state may not be sufficient for long-term hands-off object control. For example, when learning diverse jump shot sequences, it is difficult for the policy to determine how to control the ball through a single frame of HOI reference state because the ball is out of control after being shot out. One potential solution is to provide the policy with multi-frame reference states of the ball.
    \item Our method can not generalize to HOIs that are not trained on, \eg, the policy trained on \emph{back dribble} can not handle \emph{fingertip spin}.
\end{itemize}

\begin{figure*}[t]
  \centering
  \includegraphics[width=1.\linewidth]{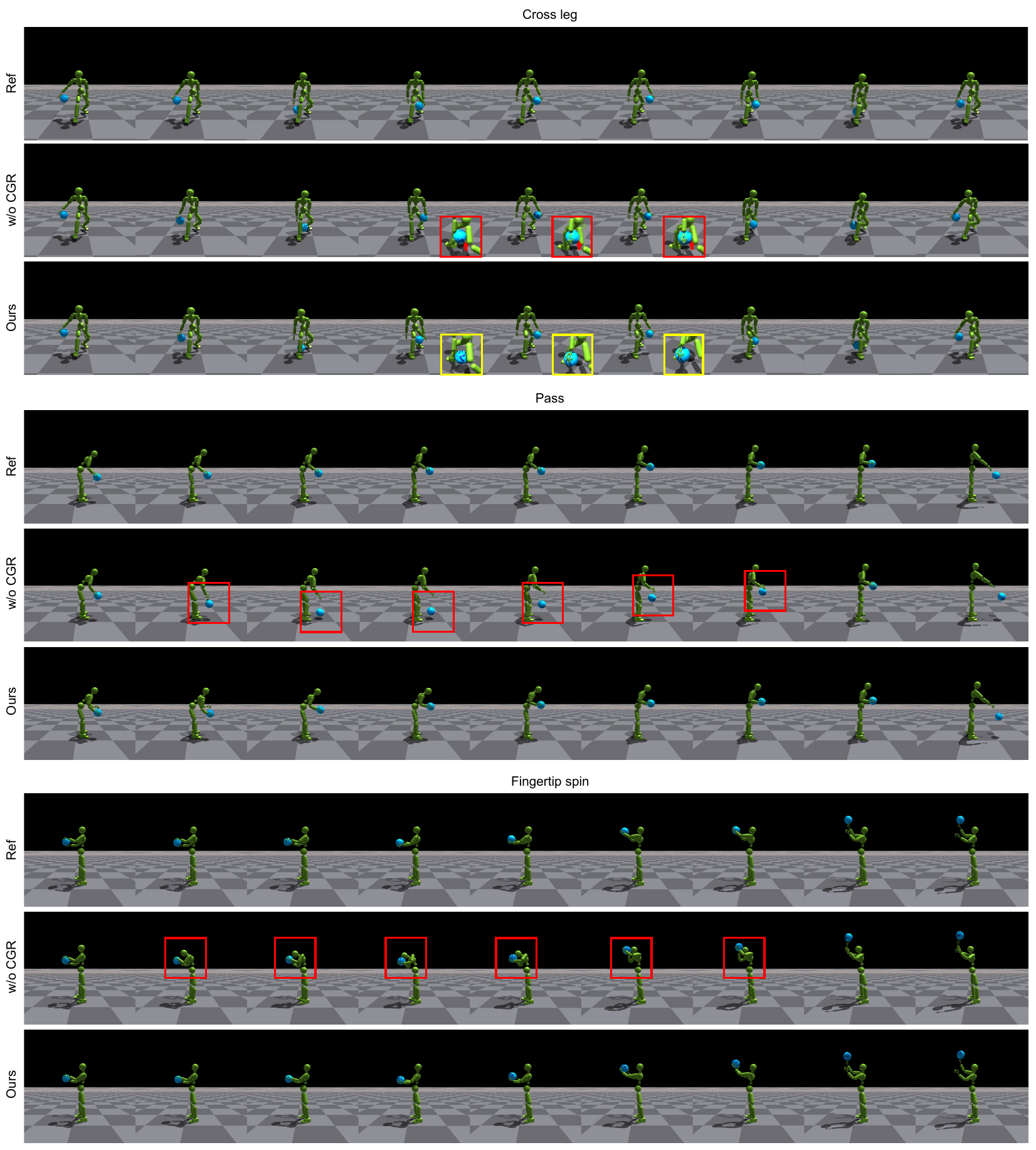}
  \caption{\textbf{Ablation on Contact Graph Reward (CGR)}. The humanoid trained w/o CGR tends to fall into the local optimal of kinematic rewards, e.g., using its left leg and left hand to hold the ball. We provide the \emph{cross leg} case another view in boxes for better observation.}
\label{fig: fullcompare} 
\end{figure*}

\subsection{Future Work}
We believe this work opens up many exciting directions for future exploration, which we list here: 
\begin{itemize}
    \item Our work demonstrates the potential to learn generic human skills from diverse HOI data. The way of acquiring HOI data accurately and conveniently is worth studying. 
    \item When large HOI data is available, studying generalization and generation based on HOI imitation will be an attractive direction toward future humanoid autonomy, considering no need for designing task-specific rewards. For example, train an MVAE \cite{mvae} using HOI imitation based on a large HOI dataset. 
    \item Although building a real humanoid with 153 DOF seems far away, it is promising to explore retargeting-based HOI Imitation, \eg, teaching a robot arm with a dextrous hand to play basketball via HOI Imitation. 
    \item Multi-human dynamic interactions are also worth studying, such as multi-player basketball games.
\end{itemize}

\begin{figure*}[t]
  \centering
  \includegraphics[width=1\linewidth]{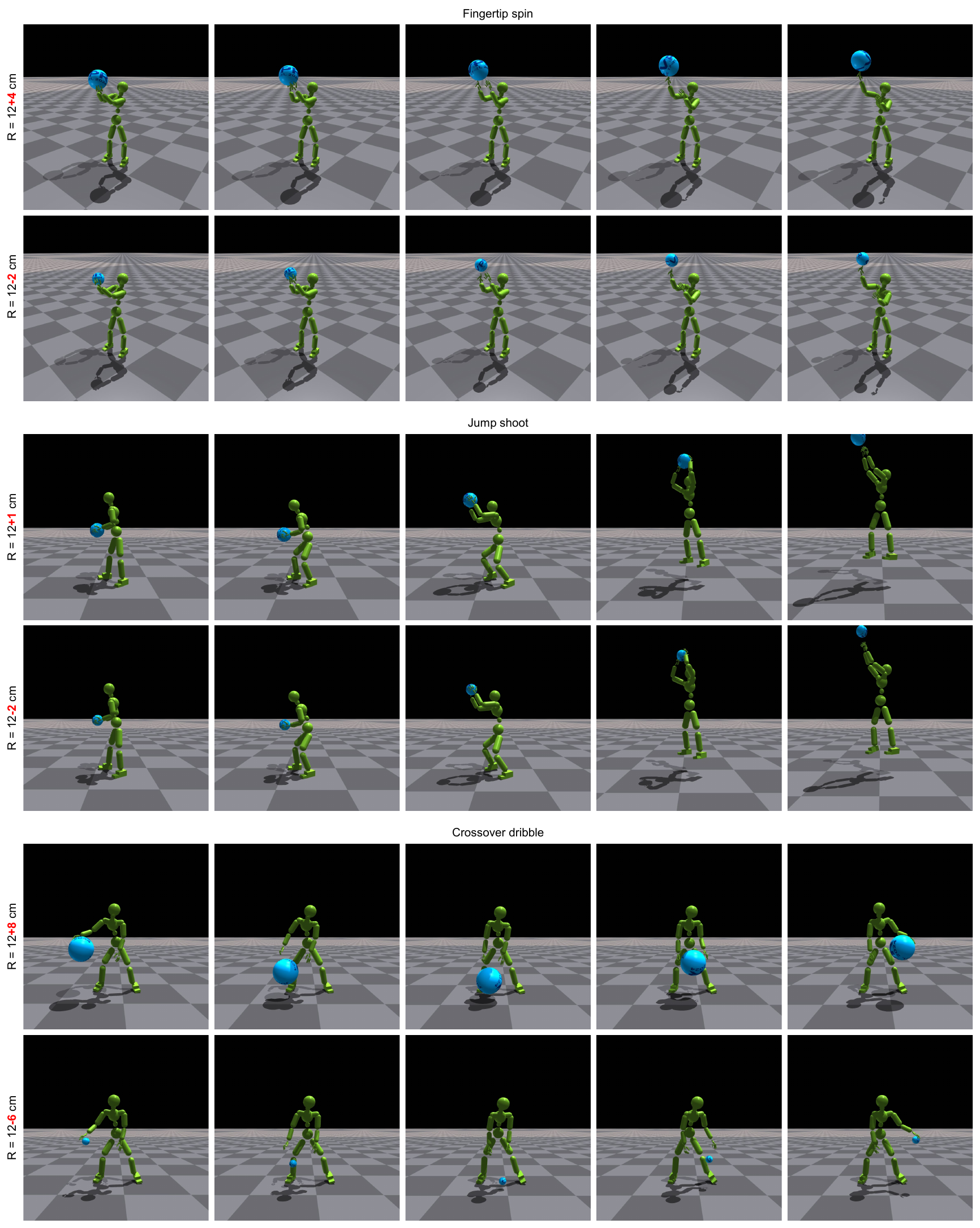}
  \caption{\textbf{Varying Ball Sizes}. During inference, we change the ball radius to different values. Interestingly, we still get reasonable results, which show the robustness of PhysHOI. The default ball radius is 12cm. R denotes the changed ball radius.}
\label{fig: ablation_radius} 
\end{figure*}

\begin{figure*}[t]
  \centering
  \includegraphics[width=1.\linewidth]{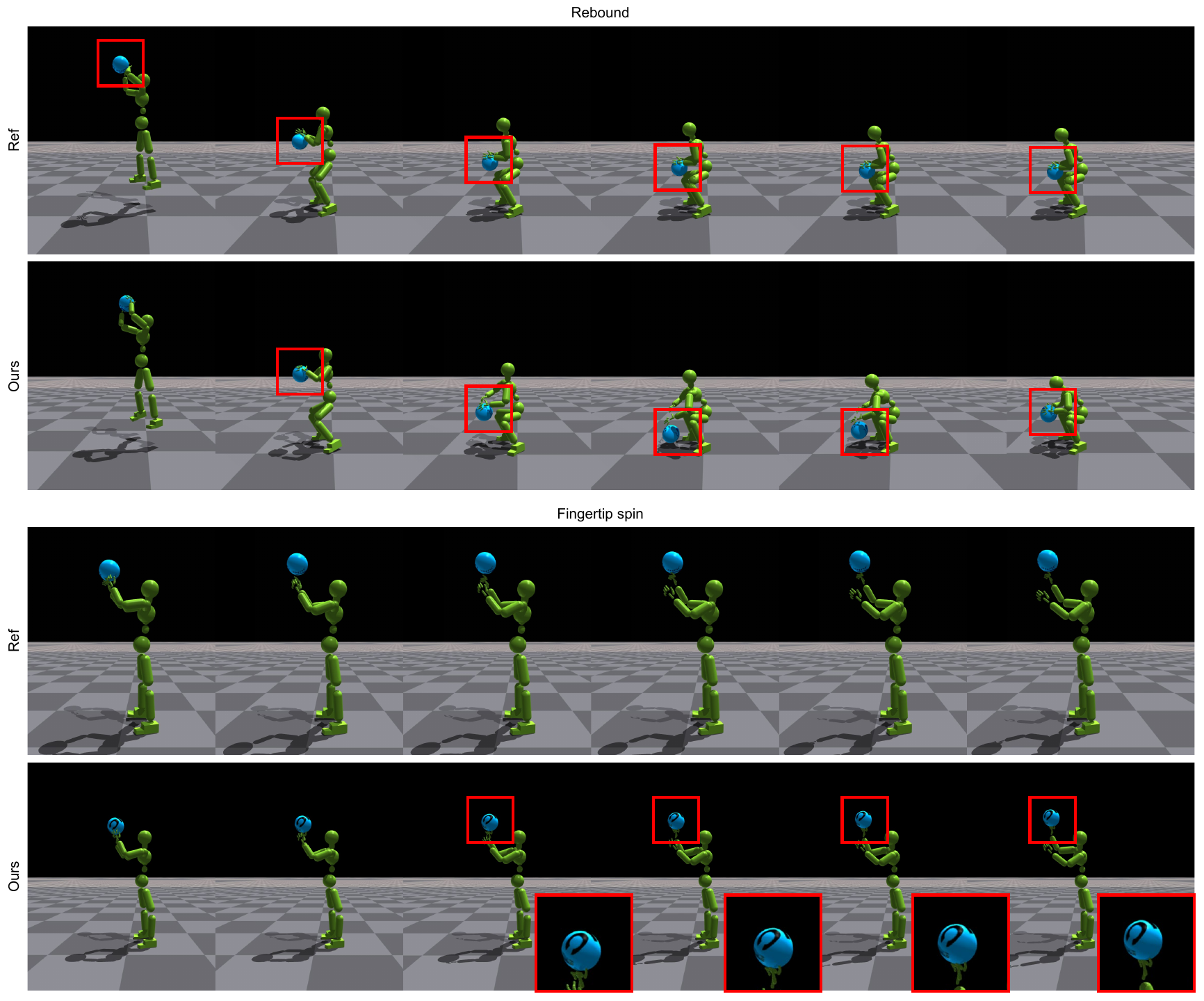}
  \caption{\textbf{Failure cases}. The failure cases of our method are mainly due to inaccurate data or incomplete contact graphs. In the \emph{rebound} case, the biased ball position results in a local optimum: the ball not being firmly grasped during the learning process. High CGR weight may lead to unpleasant interactions if the CG node is not detailed enough. In the \emph{fingertip spin} case, when we set a high CGR weight on the CG edge between hands and the ball, the trained humanoid tends to use multiple fingers to maintain steady contact. The reasons can be twofold: (1) The CG is not detailed enough, e.g., we simply take two hands as one CG node, but it is necessary to take fingers as separate CG nodes in this case. (2) The frame rate of simulation and reference data is low, which yields frequent bouncing, i.e., less contact. While reducing the corresponding CG edge weights can improve this problem, as shown in the \emph{fingertip spin} case in Fig.~\ref{fig: ablation_radius} and Fig.~\ref{fig: fullcompare}, a more general solution requires richer CGs and higher frame rates, which is also the solution we expect.}
\label{fig: failurecase} 
\end{figure*}

\end{document}